\documentclass[sigconf]{acmart}

\usepackage{booktabs}
\usepackage{setspace}
\usepackage{algpseudocode}
\usepackage{multirow}
\newtheorem*{lemma}{Lemma}
\newtheorem*{remark}{Remark}
\usepackage{makecell}
\usepackage{colortbl}

\usepackage{algorithm}
\usepackage{subcaption}

\usepackage[acronym, nohypertypes={acronym}]{glossaries}
\newacronym{model}{ImputeFormer}{\emph{Imputation Transformers}}

\AtBeginDocument{%
  \providecommand\BibTeX{{%
    \normalfont B\kern-0.5em{\scshape i\kern-0.25em b}\kern-0.8em\TeX}}}

\copyrightyear{2024}
\acmYear{2024}
\setcopyright{rightsretained}
\acmConference[KDD '24]{the 30th ACM SIGKDD Conference on Knowledge Discovery and Data Mining}{August 25--29, 2024}{Barcelona, Spain.}
\acmBooktitle{Proceedings of the 30th ACM SIGKDD Conference on Knowledge Discovery and Data Mining (KDD '24), August 25--29, 2024, Barcelona, Spain}
\acmDOI{xx}
\acmISBN{xx}

\usepackage{etoolbox}
\makeatletter
\patchcmd{\maketitle}{\@copyrightpermission}{
$^\dag$ Jian Sun is the corresponding author.\\
   \begin{minipage}{0.3\columnwidth}
     \href{https://creativecommons.org/licenses/by/4.0/}{\includegraphics[width=0.90\textwidth]{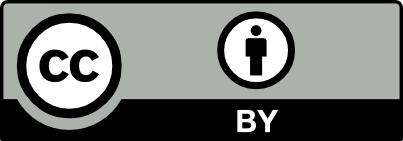}}
   \end{minipage}\hfill
   \begin{minipage}{0.7\columnwidth}
     \href{https://creativecommons.org/licenses/by/4.0/}{
     This work is licensed under a Creative Commons Attribution International 4.0 License.}
   \end{minipage}
   \vspace{5pt}
}{}{}

\makeatother

\begin{document}

\title{{ImputeFormer}: Low Rankness-Induced Transformers for Generalizable Spatiotemporal Imputation}







\author{Tong Nie$^1$,~Guoyang Qin$^1$,~Wei Ma$^{2}$,~Yuewen Mei$^1$, Jian Sun$^{1,\dag}$}
\affiliation{
  \institution{$^1$Tongji University,~~$^2$The Hong Kong Polytechnic University}
  \country{}
  \address{}
\text{nietong@tongji.edu.cn,~2015qgy@tongji.edu.cn,~wei.w.ma@polyu.edu.hk,}\\
\text{meiyuewen@tongji.edu.cn,~sunjian@tongji.edu.cn}
}

\renewcommand{\shortauthors}{Tong Nie et al.}
\renewcommand{\authors}{Tong Nie, Guoyang Qin, Wei Ma, Yuewen Mei, and Jian Sun}

\begin{abstract}
Missing data is a pervasive issue in both scientific and engineering tasks, especially for the modeling of spatiotemporal data. This problem attracts many studies to contribute to data-driven solutions. Existing imputation solutions mainly include low-rank models and deep learning models. The former assumes general structural priors but has limited model capacity. The latter possesses salient features of expressivity but lacks prior knowledge of the underlying spatiotemporal structures.
Leveraging the strengths of both two paradigms, we demonstrate a low rankness-induced Transformer to achieve a balance between strong inductive bias and high model expressivity.
The exploitation of the inherent structures of spatiotemporal data enables our model to learn balanced signal-noise representations, making it generalizable for a variety of imputation problems.
We demonstrate its superiority in terms of accuracy, efficiency, and versatility in heterogeneous datasets, including traffic flow, solar energy, smart meters, and air quality. 
Promising empirical results provide strong conviction that incorporating time series primitives, such as low-rankness, can substantially facilitate the development of a generalizable model to approach a wide range of spatiotemporal imputation problems. The model implementation is available at: \url{https://github.com/tongnie/ImputeFormer}.
\end{abstract}




\maketitle

\begin{figure*}
\centering
\includegraphics[width=1\textwidth]{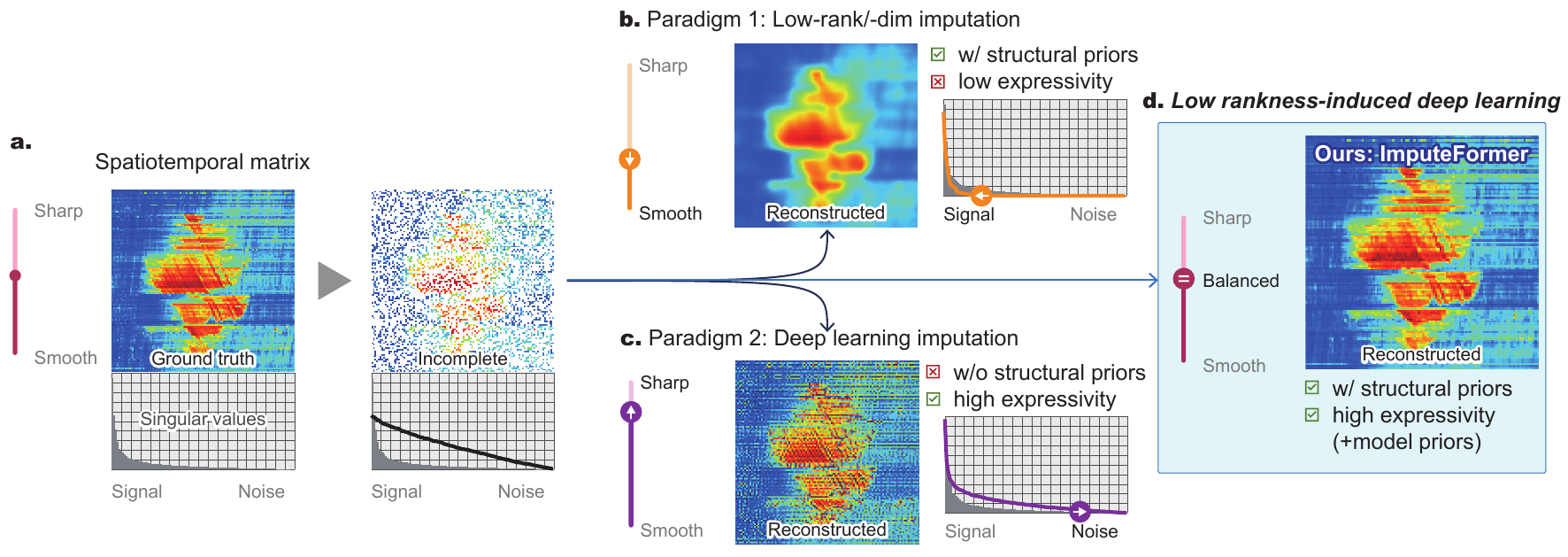}
\captionsetup{skip=1pt}
\caption{(a) The distribution of singular values in spatiotemporal data is long-tailed. The existence of missing data can increase its rank (or singular values). (b) Low-rank models can filter out informative signals and generate a smooth reconstruction, resulting in truncating too much energy in the left part of its spectrum. (c) Deep models can preserve high-frequency noise and generate sharp imputations, maintaining too much energy for the right part of the singular spectrum.
With the generality of low-rank models and the expressivity of deep models, \texttt{\acrshort{model}} achieves a signal-noise balance for accurate imputation.
}
\label{FIG:motivation}
\end{figure*}
\section{Introduction}
Missing data is a common challenge in detection systems, especially in high-resolution monitoring systems. Factors such as inclement weather, energy supply, and sensor service time can adversely affect the quality of monitoring data \cite{chen2021bayesian}. Given these factors, data missing rates can be quite high. 
For example, the air quality measurements in the Urban Air project \cite{zheng2015forecasting} contain about 30\% invalid records due to station malfunction.
Similarly, subsets of Uber Movement data, such as New York City, have an approximate 85\% of missing records. This problem encourages researchers to develop advanced models that can exploit limited observations to impute missing values. Extensive research has contributed to data-driven methods for this purpose, especially in the field of spatiotemporal data \cite{chen2020nonconvex,ye2021spatial,chen2021bayesian,nie2022truncated,nie2023correlating,wang2023low}.

Generally, there are two research paradigms for imputing missing data. The first paradigm uses low-rank and low-dimensional analytical models, such as \cite{chen2020nonconvex,chen2021low,nie2022truncated,liu2022multivariate,wang2023low}, which assumes the data has a well-structured matrix or tensor form. These models utilize the algebraic properties of the assumed structure, such as low-rankness, low nuclear norm, and spectrum sparsity (we use the term ``low-rank'' as a proxy), to impute missing values \cite{ma2019missing}. While simple low-rank models like matrix factorization and tensor completion can effectively handle incomplete data, they may struggle with capturing complex patterns, such as nonlinearity and nonstationarity. 
With strong inductive bias, these models can excessively smooth the reconstructed data, filtering out informative signals, and generating oversmoothing reconstructions in some cases \cite{muhammad2018image}.

The second paradigm uses deep learning-based imputation models. These models learn the dynamics of the data-generating process and demonstrate improved performance \cite{cao2018brits,che2018recurrent,GRIN,du2023saits,SPIN,liu2023pristi}. However, despite the success of these models on various benchmarks, there are still costs and difficulties that need further attention. 
First, such data-intensive methods require a substantial amount of training expenses due to the complex model structures in deep learning, such as probabilistic diffusion and bidirectional recurrence \cite{cao2018brits,tashiro2021csdi}. This can consume significant computational memory and resources, making them less efficient for real-time deployment.
Second, empirical loss-based learning methods, without the guidance of physics or data structures, are prone to overfitting and perform poorly when applied to tasks fall outside of the distribution of training data \cite{zhang2020physics}.

With the shift of focus in the field of deep imputation from RNNs and diffusion models to Transformers \cite{ma2019cdsa,du2023saits,SPIN}, Transformer-related architectures have gained significant attention due to their potential to provide efficient generative output and high expressivity, enabling more effective imputations compared to autoregression-based models.
Additionally, Transformers are considered foundational architectures for general time series forecasting \cite{TimeGPT}. 
However, the effectiveness of applying Transformers to general data imputation tasks requires further investigation.
Modern deep learning techniques associated with these architectures, such as self-attention and residual connections, can unintentionally preserve high-frequency noise in data as informative signals \cite{sharma2023truth}. This can lead the model to learn high-rank representations that violate the natural distribution of data.
Furthermore, the existence of missing data can introduce spurious correlations between ``tokens,'' posing challenges to these architectures. 
Considering the above-mentioned concerns, incorporating a low-rank inductive bias into the Transformer framework seems to provide a chance to improve both the effectiveness and efficiency in spatiotemporal imputation.

In summary, matrix- and tensor-based models offer useful priors for spatiotemporal data, such as low-rankness and sparsity. However, their ability to represent data is limited (see Fig. \ref{FIG:motivation}(b)). On the other hand, deep learning models, particularly Transformers, excel at learning representations but lack prior knowledge of data generation (see Fig. \ref{FIG:motivation}(c)). As the demand for a versatile and adaptable model that can handle various imputation problems in reality increases, such as cross-domain datasets, different observation conditions, highly sparse measurements, and different input patterns, it becomes apparent that existing advanced solutions, typically evaluated on limited tasks with simple settings, may not be generalizable.
Hence, there is a temptation to merge these two paradigms and utilize their respective strengths to investigate an alternative paradigm that can effectively handle complex imputation scenarios.

%
To this end, in this paper we leverage the structural priors of low-rankness to generalize the canonical Transformer (see Fig. \ref{FIG:motivation}(d)) in general spatiotemporal imputation tasks. Our approach, referred to as \gls{model}, imposes low-rankness and achieves attention factorization equivalently by introducing a projected attention mechanism on the temporal dimension and an embedded attention on the spatial dimension. Additionally, we propose a Fourier sparsity loss to regularize the solution's spectrum. 
By inheriting the merits of both low-rank and deep learning models, it has achieved state-of-the-art imputation performance on various benchmarks.
Our main contributions are summarized as follows:
\begin{enumerate}
    \item We are among the first to empower Transformers with low-rankness inductive bias to achieve a balance between signal and noise for general spatiotemporal data imputation;
    \item Compared to state-of-the-art benchmark models, we demonstrate the advantages of \texttt{\acrshort{model}} in accuracy, efficiency, and versatility in diverse datasets, such as traffic flow, solar energy, electricity consumption, and air quality;
    \item Comprehensive case studies reveal the model's interpretability and provide insights into the deep imputation paradigm.
\end{enumerate}

\begin{figure*}[!ht]
\centering
\includegraphics[width=1\textwidth]{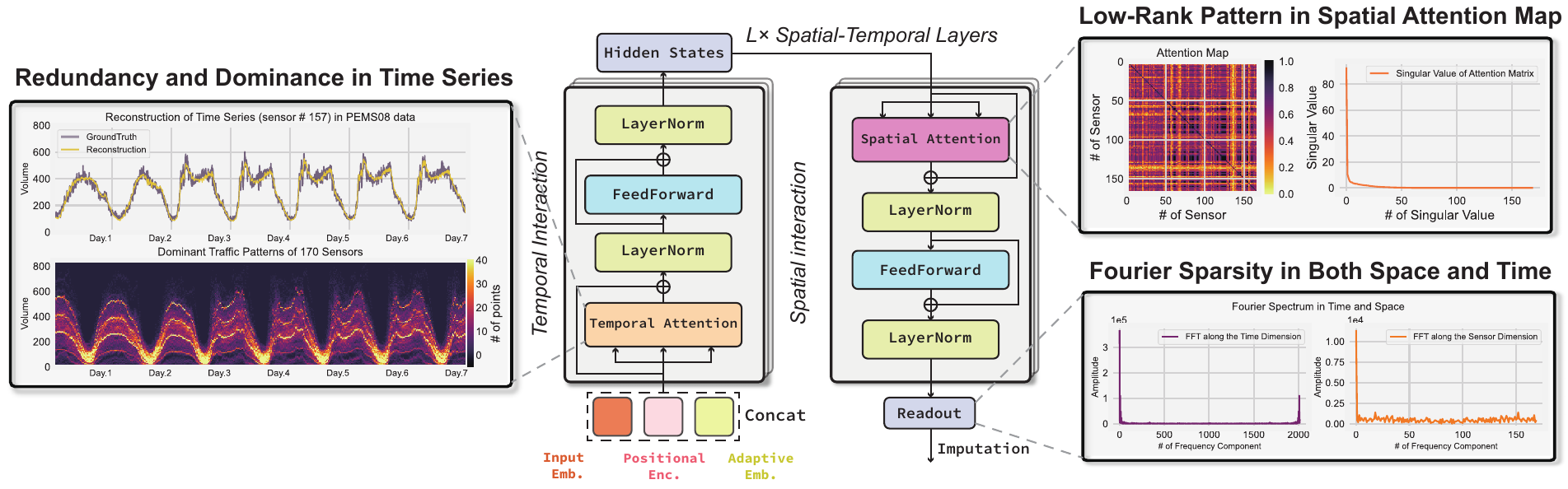}
\captionsetup{skip=1pt}
\caption{Low-rankness in time series and the induced \texttt{\acrshort{model}}. 
(a) Redundancy in time series: PEMS08 data can be reasonably reconstructed using only five dominant patterns.
(b) Low-rank spatial attention map: the singular values of the multivariate attention map show a long-tailed distribution and most of them are small values. 
(c) Fourier sparsity in both space and time axes: both the spatial and temporal signals possess a sparse Fourier spectrum, with most amplitudes close to zero. 
}
\label{FIG:intro}
\end{figure*}
\vspace{-0.2cm}
\section{Preliminary}\label{sec:preliminary}
\noindent\textbf{{Notations}}. This section first introduces some basic notations following \cite{SPIN}. 
In a continuously working sensor system with $N$ static detectors at some measurement positions, spatiotemporal data with context information can be obtained: (1) $\mathbf{X}_{t:t+T}\in\mathbb{R}^{N\times T}$: The observed data matrix containing missing values collected by all sensors over a time interval $\mathcal{T}=\{t,\dots,t+T\}$, where $T$ represents the observation period; (2) $\mathbf{Y}_{t:t+T}\in\mathbb{R}^{N\times T}$: The ground truth data matrix used for evaluation; (3) ${\mathbf{U}_{t:t+T} \in \mathbb{R}^{T \times d_\text{u}}}$: Exogenous variables describe time series, such as the time of day, day of week, and week of month information; (4) ${\mathbf{V} \in \mathbb{R}^{N\times d_\text{v}}}$: Meta information of sensors, such as detector ID and location of installation.

\noindent\textbf{{Problem Formulation}}.
The multivariate time series imputation problem defines an inductive learning and inference process: 
\begin{equation}\label{eq:formulation}
\begin{aligned}
&\texttt{Learning}~\widehat{\Theta}=\arg\min_{\Theta}\sum_{t\in\mathcal{T}}\ell\left(\texttt{NN}(\{\mathbf{x}_{t},\mathbf{u}_{t},\mathbf{m}_{t}\},\mathbf{V}|\Theta),\mathbf{x}_{t}\right), \\
&\texttt{Inference}~\widehat{\mathbf{x}}_{t'}=\texttt{NN}(\{\mathbf{x}_{t'},\mathbf{u}_{t'},\mathbf{m}_{t'}\},\mathbf{V}|\widehat{\Theta}), \forall\{t'\dots t'+T\},\\
\end{aligned}
\end{equation}
where $\texttt{NN}(\cdot|\Theta)$ is the neural network model parameterized by $\Theta$, and the indicator $\mathbf{m}_{t}$ denotes the locations of the masked values for training and the locations of the observed values for inference. After training the model on observed data, the imputation model can act in a different time span than the training set.

\vspace{-0.2cm}
\section{Related Work}
Generally, there exist two series of studies on multivariate time series imputation, i.e., 1) low-dimensional/rank models and 2) deep imputation models. 
We particularly discuss existing Transformer-based solutions to clarify the connections between our models. 

\noindent\textbf{Low-Dimensional/Rank Imputation}. Early methods addressed the data imputation problem by exploring statistical interpolation tools, such as \texttt{MICE} \cite{van2011mice}. Recently, low-rank matrix factorization \cite{yu2016temporal, liu2022multivariate} and tensor completion \cite{chen2020nonconvex,chen2021low, chen2021bayesian, nie2022truncated, nie2023correlating} have emerged as numerically efficient techniques for spatiotemporal imputation. To incorporate series-related features, \texttt{TRMF} \cite{yu2016temporal} imposed autoregressive regularization on the temporal manifold.
\texttt{TiDER} \cite{liu2022multivariate} decomposed the time series into trend, seasonality and bias components under the factorization framework. Despite being conceptually intuitive and concise, limited capacity hinder their practical effectiveness.

\noindent\textbf{Deep Learning Imputation}. Recent advances in neural time series analysis open a new horizon to improve imputation performance. 
Generally, deep imputation methods learn to reconstruct the distribution of observed data or aggregate pointwise information progressively \cite{GNN4TS}. Representative methods include \texttt{GRU-D} \cite{che2018recurrent}, \texttt{GRUI} \cite{luo2018multivariate}, \texttt{BRITS} \cite{cao2018brits}, \texttt{GAIN} \cite{yoon2018gain}, \texttt{E2GAN} \cite{luo2019e2gan}  \texttt{NAOMI} \cite{liu2019naomi},   \texttt{CSDI} \cite{tashiro2021csdi}, and \texttt{PriSTI} \cite{liu2023pristi}.
 To exploit the multivariate nature of spatiotemporal data, graph neural networks (GNNs) have been adopted to model sensor-wise correlations for more complicated missing patterns.
For example, \texttt{MDGCN} \cite{liang2022memory} and \texttt{GACN} \cite{ye2021spatial} applied GNNs with RNNs for traffic data imputation. \texttt{IGNNK} \cite{wu2021inductive}, \texttt{STAR} \cite{liang2023spatial}, and \texttt{STCAGCN} \cite{nie2023towards} further tackle the kriging problem, which is a special data imputation scenario. As a SOTA model and an architectural template for GNN-RNNs, \texttt{GRIN} \cite{GRIN} based on message-passing GRUs that progressively performed a two-stage forward and backward recurrent message aggregation with predefined relational biases.

\noindent\textbf{Transformers for Time Series Imputation}. Transformers \cite{transformer} can aggregate abundant information from arbitrary input elements, becoming a natural choice for sequential data imputation. In particular, \texttt{CDSA} \cite{ma2019cdsa} developed a cross-channel attention that utilizes correlations in different dimensions.
\texttt{SAITS} \cite{du2023saits} combined the masked imputation task with an observed reconstruction task, and applied a diagonally-masked self-attention to hierarchically reconstruct sparse data.
\texttt{SPIN} \cite{SPIN} achieved SOTA imputation performance by conducting a sparse cross-attention and a temporal self-attention on all observed spatiotemporal points. 
However, the high complexity of cross-attention hinders its application in larger graphs.
%


\vspace{-0.2cm}
\section{Low Rankness-Induced Transformer}\label{methodology}
This section elaborates the \texttt{\acrshort{model}} model. 
The major difference between our model and the canonical Transformer is the integration of low-rank factorization. Unlike GNNs, \texttt{\acrshort{model}} does not require a predefined graph due to the global adaptive interaction between series. It also bypasses the use of intricate temporal techniques, such as bidirectional recurrent aggregation, sparse cross-attention, and attention masking. Furthermore, it achieves linear complexity with respect to spatial and temporal dimensions.

\subsection{Architectural Overview}\label{subsec:overview}
The overall structure of the proposed \texttt{\acrshort{model}} is shown in Fig. \ref{FIG:intro}. The input embedding layer projects sparse observations to hidden states in an additional dimension and introduces both fixed and learnable embedding into the inputs. 
Following a time-and-graph template \cite{Taming}, {TemporalInteraction} and {SpatialInteraction} perform global message passing alternatively at all spatiotemporal coordinates. Finally, a $\texttt{MLP}$ readout is adopted to output the final imputation. This process can be summarized as follows:
\begin{equation}
\begin{aligned}
    \mathcal{Z}_{t:t+T}^{(0)}=&\texttt{InputEmb}(\mathbf{X}_{t:t+T},\mathbf{U}_{t:t+T},\mathbf{V}), \\
    \mathcal{Z}_{t:t+T}^{(\ell+1)}=&\texttt{TemporalInteraction}(\mathcal{Z}_{t:t+T}^{(\ell)}),\\
    \mathcal{Z}_{t:t+T}^{(\ell+1)}=&\texttt{SpatialInteraction}(\mathcal{Z}_{t:t+T}^{(\ell+1)}), \forall \ell\in\{0,\dots L\},\\
    \widehat{\mathbf{X}}_{t:t+T}=&\texttt{Readout}(\mathcal{Z}_{t:t+T}^{(L+1)}).
\end{aligned}
\end{equation}

The canonical Transformer block \cite{transformer} can be adopted to gather spatial-temporal information for imputation. 
However, we argue that directly applying it to the imputation problem is questionable, and there exist three key concerns: (1) \textit{Spurious correlations}: Short-term series within a window can be noisy and indistinguishable. Modeling relational structures using sparse input can cause spurious and misleading correlations. 
(2) \textit{High-rank estimations}: Time series are typically low-rank in nature \cite{liu2022recovery}. Full-attention computation on raw data can be overcorrelated and generate high-rank attention maps.
(3) \textit{Scalability issue}: All pairwise attention on large graphs is memory intensive and computationally inefficient. To address these issues, we start from time series primitives and enhance the Transformer using these structural priors.

\subsection{Spatiotemporal Input Embedding}\label{sec:input}
\noindent\textbf{{Input Embedding}}.
We adopt a dimension expansion strategy \cite{MTGNN} to preserve the information density of the incomplete time series. In practice, we expand an additional dimension of the input and project it into a hidden state along this new dimension:
\begin{equation}\label{input_unsqueeze}
    \mathcal{Z}_{t:t+T}^{(0)}=\texttt{MLP}(\texttt{Unsqueeze}(\mathbf{X}_{t:t+T}, \text{dim=-1}),
\end{equation}
where $\texttt{Unsqueeze}(\cdot):\mathbb{R}^{N\times T}\rightarrow\mathbb{R}^{N\times T\times 1}$, and $\mathcal{Z}_{t:t+T}^{(0)}\in\mathbb{R}^{N\times T\times D}$ is the initial hidden representation. With this, we can aggregate message from other time points by learning data-dependent weights:
\begin{equation}\label{time_varying_ar}
\mathbf{Z}_{t:t+T}^{i,(\ell+1)}=\mathcal{F}_\ell(\mathbf{Z}_{t:t+T}^{i,(\ell)})\mathbf{Z}_{t:t+T}^{i,(\ell)},
\end{equation}
\noindent where $\mathcal{F}_\ell(\cdot):\mathbb{R}^{T\times D}\rightarrow\mathbb{R}^{T\times T}$ represents a data-driven function at the $\ell$-th layer, such as self-attention. The rationale of this strategy is discussed in \ref{appendix: input_embedding}.

\noindent\textbf{Time Stamp Encoding}.
Time stamp encoding is adopted to handle the order-agnostic nature of Transformers \cite{transformer}.
As the input series covers a relatively short range, we only consider the time-of-day information. We adopt the sinusoidal positional encoding in \cite{transformer} to inject the time-of-day information of each time series:
\begin{equation}
    \begin{aligned}
        p_{\text{sine}}^t&=\texttt{sin}(p_t*2\pi/\delta_D), ~
        p_{\text{cosine}}^t=\texttt{cos}(p_t*2\pi/\delta_D),\\
        \mathbf{u}_t &= \left[p_{\text{sine}}^t \| p_{\text{cosine}}^t\right],
    \end{aligned}
\end{equation}
where $p_t$ is the index of $t$-th time-of-day point in the series, and $\delta_D$ is the day-unit time mapping. We concatenate $\mathbf{p}_{\text{sine}}$ and $\mathbf{p}_{\text{cosine}}$ as the final time stamp encoding $\mathbf{U}_{t:t+T}\in\mathbb{R}^{ T\times 2}$. 

\noindent\textbf{{Node Embedding}}.
Previous work has demonstrated the importance of node identification in distinguishing different sensors for spatiotemporal forecasting \cite{STID,Taming,nie2023contextualizing, STAEformer}. Here we also recommend the use of learnable node embedding for imputation task. On the one hand, it benefits the adaptation of local components \cite{Taming} in graph-based data structure. On the other hand, we highlight that node embedding can be treated as an abstract and low-dimensional representation of the incomplete series.
To implement, we assign each series a randomly initialized parameter $\mathbf{e}^i\in\mathbb{R}^{D_s}$. We then split the hidden dimension of the static node embedding equally by the length of the time window 
as a \textit{multi-head} node embedding and unfold it to form a low-dimensional and time-varying representation: $\mathbf{E}^i_{t:t+T}\in\mathbb{R}^{T \times D_s/T}$. Implicit interactions between node embedding, input data, and modular components are involved in the end-to-end gradient descent process. Finally, the spatiotemporal input embedding for each node can be formulated as follows:
\begin{equation}\label{eq:input_embed}
    \mathbf{Z}_{t:t+T}^{i,(1)}=\texttt{Concat}(\mathbf{Z}_{t:t+T}^{i,(0)}; \mathbf{U}_{t:t+T} ;\mathbf{E}^i_{t:t+T},\text{dim=-1}),
\end{equation}
where $\mathbf{Z}_{t:t+T}^{i,(1)}\in\mathbb{R}^{T\times (D+ D_s/T+2)}$ is input to the following modules. 

\subsection{Temporal Projected Attention}\label{subsec:temporal}

As is evident in Fig. \ref{FIG:intro}, time series are supposed to be redundant in the time domain, that is, most of the information can be reconstructed using only a few dominant modes. However, as the hidden dimension $D'$ is practically much larger than the sequence length $T$, the attention score $\mathbb{R}^{T\times D'}\times\mathbb{R}^{D'\times T}\rightarrow\mathbb{R}^{T\times T}$ can be a high-rank matrix, which is both adverse and inefficient to reconstruct incomplete hidden spaces. 
To address this concern, we propose a new projected attention mechanism to impose a low-rank constraint on the attentive process and efficiently model pairwise temporal interactions between time points in linear complexity. 

To utilize this structural bias, we first project the initial features to dense representations by attending to a low-dimensional vector. Specifically, we first randomly initialize a learnable vector that is shared by all nodes with the gradient tractable as the \textit{projector} $\mathbf{P}_{\text{proj}}\in\mathbb{R}^{C\times D'}$, where $C<T$ is the projected dimension. In order to represent the temporal message in a compact form, we then project the hidden states $\mathbf{Z}^{i, (\ell)}\in\mathbb{R}^{T\times D'}$ (subscripts are omitted for brevity) to the projected space by attending to the query projector:
\begin{equation}\label{proj_att}
\begin{aligned}
    \widetilde{\mathbf{Z}}_{\text{proj}}^{i,(\ell)}&=\texttt{SelfAtten}(\mathbf{P}_{\text{proj}}^{(\ell)},\mathbf{Z}^{i,(\ell)},\mathbf{Z}^{i,(\ell)})),\\
    &=\texttt{Softmax}\left(\frac{\mathbf{P}_{\text{proj}}^{(\ell)}\mathbf{W}_{\text{Q}}\mathbf{W}_{\text{K}}^{\mathsf{T}}\mathbf{Z}^{i,(\ell),\mathsf{T}}}{\sqrt{D'}}\right)\mathbf{Z}^{i,(\ell)}\mathbf{W}_{\text{V}},\\
\end{aligned}
\end{equation}
where $\widetilde{\mathbf{Z}}_{\text{proj}}^{i,(\ell)}\in\mathbb{R}^{C\times D'}$ is the projected value, $\mathbf{W}_{\text{Q}},\mathbf{W}_{\text{K}},\mathbf{W}_{\text{V}}\in\mathbb{R}^{D'\times D'}$ are linear weights. In particular, since the projector $\mathbf{P}_{\text{proj}}$ is decoupled from the spatial dimension, the resulted attention map $\texttt{Softmax}(\mathbf{P}_{\text{proj}}^{(\ell)}\mathbf{W}_{\text{Q}}\mathbf{W}_{\text{K}}^{\mathsf{T}}\mathbf{Z}^{i,(\ell),\mathsf{T}}/{\sqrt{D'}})\in\mathbb{R}^{C\times T}$ can be interpreted as an indicator of how the incomplete information flow can be compressed into a compact representation with smaller dimension, that is, an aggregation of available messages. More expositions on the projector will be provided in Section \ref{case_study:projector}.

$\widetilde{\mathbf{Z}}_{\text{proj}}^{i,(\ell)}$ stores the principal temporal patterns within the input data. Then, we can recover the complete series with this compact representation by dispersing the projected information to all other full-length series by using the projector as a key dictionary:
\begin{equation}\label{proj_att_all}
\begin{aligned}
    \mathbf{Z}^{i,(\ell)}_{\text{hat}}&=\texttt{SelfAtten}(\mathbf{Z}^{i,(\ell)},\mathbf{P}_{\text{proj}}^{(\ell)},\widetilde{\mathbf{Z}}_{\text{proj}}^{i,(\ell)})),\\
    &=\texttt{Softmax}\left(\frac{\mathbf{Z}^{i,(\ell)}\mathbf{W}_{\text{Q}}\mathbf{W}_{\text{K}}^{\mathsf{T}}\mathbf{P}_{\text{proj}}^{(\ell),\mathsf{T}}}{\sqrt{D'}}\right)\widetilde{\mathbf{Z}}_{\text{proj}}^{i,(\ell)}\mathbf{W}_{\text{V}},\\
\end{aligned}
\end{equation}
then the above process is integrated into the Transformer encoder:
\begin{equation}
\begin{aligned}
    \widehat{\mathbf{Z}}^{i,(\ell)}&=\texttt{LayerNorm}(\mathbf{Z}^{i,(\ell)}+\mathbf{Z}^{i,(\ell)}_{\text{hat}}), \\
    \mathbf{Z}^{i,(\ell+1)}&=\texttt{LayerNorm}(\widehat{\mathbf{Z}}^{i,(\ell)}+\texttt{FeedForward}(\widehat{\mathbf{Z}}^{i,(\ell)})), \\
\end{aligned}
\end{equation}
where $\mathbf{Z}^{i,(\ell+1)}\in\mathbb{R}^{T \times D'}$ is the imputation by the $\ell$-th temporal interaction layer. Since the projector in Eqs. \eqref{proj_att} and \eqref{proj_att_all} can be obtained by end-to-end learning and is independent of the order in series, it has the property of data-dependent model in Eq. \eqref{time_varying_ar}. To indicate how the above process learns the low-rank representation of temporal attention, we develop the following remark.

\begin{remark}[Difference between projected attention and canonical self-attention] Given the query-key-value matrix $\mathbf{Q},\mathbf{K},\mathbf{V}\in\mathbb{R}^{T \times D'}$, the canonical self-attention \cite{transformer} in the temporal axis can be expressed compactly as: $\texttt{SelfAtten}(\mathbf{Q},\mathbf{K},\mathbf{V})=\sigma(\mathbf{Q}\mathbf{K}^{\mathsf{T}})\mathbf{V}$ with the rank $r\leq\min\{T,D'\}$. 
For comparison, the two-step attentive process in Eqs. \eqref{proj_att} and \eqref{proj_att_all} are equivalent to the expanded form:
\begin{equation*}
\begin{aligned}
    \widehat{\mathbf{Z}}&=\texttt{SelfAtten}(\mathbf{Q},\mathbf{P},\texttt{SelfAtten}(\mathbf{P},\mathbf{K},\mathbf{V})), \\
    &=\sigma(\mathbf{Q}\mathbf{P}^{\mathsf{T}})\texttt{SelfAtten}(\mathbf{P},\mathbf{K},\mathbf{V}),\\
    &=\sigma(\mathbf{Q}\mathbf{P}^{\mathsf{T}})\sigma(\mathbf{P}\mathbf{K}^{\mathsf{T}})\mathbf{V} \approx \frac{1}{N^2}\mathbf{Q}(\mathbf{P}^{\mathsf{T}}\mathbf{P})\mathbf{K}^{\mathsf{T}}\mathbf{V}.
\end{aligned}
\end{equation*}
Recall that the projector $\mathbf{P}\in\mathbb{R}^{C\times D'}$ can have a small projection dimension $C$, it can be viewed as a channel-wise matrix factorization to reduce redundancy within each time series. The rank of the projected attention matrix is $r\leq\min\{C,D'\}$, which is theoretically lower than the original rank. The projected attention guarantees expressivity by maintaining a large hidden dimension $D'$, while at the same time admitting a low-rank solution using a small projection dimension $C$.
The rank-reduced temporal attention matrix exploits the low-rankness of data in the temporal dimension, which is different from the low-rank adaptation of model parameters developed recently \cite{sharma2023truth}.
\end{remark}

The ``projection-reconstruction" process in Eqs. \eqref{proj_att} and \eqref{proj_att_all} resemble the low-rank factorization process $\mathbf{X}=\mathbf{U}\mathbf{V}^{\mathsf{T}}$. Inflow in Eq. \eqref{proj_att} controls the amount of message used to form a dense representation in a lower-dimensional space. Outflow in Eq. \eqref{proj_att_all} determines how hidden states can be reconstructed using only a few projected coordinates. 
This mechanism also brings about efficiency benefits. The canonical self-attention costs $\mathcal{O}(T^2)$ time complexity. 
The complexity of the projected attention is $\mathcal{O}(TC)$, which scales linearly (see Section \ref{appendix: low-rank}) and is efficient for longer sequences. 
In addition, low-rank attention preserves the dominating correlational structures and eliminates spurious correlations. The cleaned correlations allow the model to focus on the most relevant data as a reference.

\subsection{Spatial Embedded Attention}\label{sec:graph_conv}
The availability of observed temporal information is not sufficient for fine-grained imputation. In some cases, specific spatial events, such as traffic congestion, can lead to non-local patterns and unusual records.
Therefore, it is reasonable to exploit the multivariate relationships between series as a complement.
A straightforward way to address this problem is to apply a Transformer block in spatial dimension \cite{STAEformer,ye2021spatial}. Nevertheless, the three concerns discussed in Section \ref{subsec:overview} prevent the direct use of this technique.

Consequently, we design an embedded attention as an alternative to spatial attention. 
We highlight that the node embedding in Eq. \eqref{eq:input_embed} signifies not only the identity of series, but also a dense abstract of each individual. We then establish a correlation map using this \textit{low-dimensional agent}. Formally, we assume that the message passing happens on a fully connected dense graph, and the edge weights are estimated by the pairwise correlating of node embedding:
\begin{equation}\label{eq:graph_conv}
\begin{aligned}
\mathbf{Q}_e^{(\ell)}&= \texttt{Linear}(\mathbf{E}), ~\mathbf{K}_e^{(\ell)}= \texttt{Linear}(\mathbf{E}),\\
\mathbf{A}^{(\ell)}&=\texttt{Softmax}\left(\frac{\mathbf{Q}_e^{(\ell)}\mathbf{K}_e^{(\ell)\mathsf{T}}}{\sqrt{D'}}\right), \\
\end{aligned}
\end{equation}
where $\mathbf{A}^{(\ell)}\in\mathbb{R}^{N\times N}$ denotes the pairwise correlation score of all sensors, and $\mathbf{Q}_e^{(\ell)},\mathbf{K}_e^{(\ell)}\in\mathbb{R}^{N\times D_{\text{emb}}}$ are linearly projected from the spatiotemporal embedding set $\mathbf{E}=[\Bar{\mathbf{e}}^1\|\Bar{\mathbf{e}}^2\|\cdots\|\Bar{\mathbf{e}}^N]\in\mathbb{R}^{N\times D_{s}/T}$, with $\Bar{\mathbf{e}}^i$ being the static node embedding averaging over the temporal heads $\{t:t+T\}$ from Eq. \eqref{eq:input_embed}.

Given the graph representation over a period $\mathcal{Z}\in\mathbb{R}^{N\times T \times D'}$, the complexity of obtaining a full spatial attention matrix costs $\mathcal{O}(N^2TD')$.
To alleviate scalability concerns on large graphs, we adopt the normalization trick in \cite{shen2021efficient} to reparameterize Eq. \eqref{eq:graph_conv}. Observe that the main bottleneck in Eq. \eqref{eq:graph_conv} happens in the multiplication of two large matrix $\mathbf{Q}_e^{(\ell)}$ and $\mathbf{K}_e^{(\ell)}$, we can reduce the complexity using the associative property of matrix multiplication if we can decouple the \texttt{softmax} function. To this end, we apply the \texttt{softmax} on separate side of the Q-K matrix and approximate $\mathbf{A}$ as:
\begin{equation}\label{eq:efficient_att}
    \mathbf{A}^{(\ell)}\approx\sigma_2(\widetilde{\mathbf{Q}}_e^{(\ell)})\sigma_1(\widetilde{\mathbf{K}}_e^{(\ell)})^{\mathsf{T}},
\end{equation}
where $\sigma(\cdot)$ is the abbreviation for \texttt{softmax}, and the subscript denotes the dimension that we perform normalization. The scaled Q-K functions $\widetilde{\mathbf{Q}}_e^{(\ell)}=\mathbf{Q}_e^{(\ell)}/{\|\mathbf{Q}_e^{(\ell)}\|_F}$, $\widetilde{\mathbf{K}}_e^{(\ell)}=\mathbf{K}_e^{(\ell)}/{\|\mathbf{K}_e^{(\ell)}\|_F}$ are used to ensure numerical stability. On top of Eq. \eqref{eq:efficient_att}, the complete embedded attention can be reformulated as:
\begin{equation}\label{eq:graph_conv_efficient}
\begin{aligned}
    \widetilde{\mathbf{Z}}^{(\ell)}_t&=\texttt{LayerNorm}(\mathbf{Z}_t^{(\ell)}+ \sigma_2(\widetilde{\mathbf{Q}}_e^{(\ell)})\sigma_1(\widetilde{\mathbf{K}}_e^{(\ell)})^{\mathsf{T}}\mathbf{Z}^{(\ell)}_t), \\
   &=\texttt{LayerNorm}(\mathbf{Z}_t^{(\ell)}+\sigma_2(\widetilde{\mathbf{Q}}_e^{(\ell)})\left(\sigma_1(\widetilde{\mathbf{K}}_e^{(\ell)})^{\mathsf{T}}\mathbf{Z}^{(\ell)}_t\right)), \\
   \mathbf{Z}_t^{(\ell+1)}&=\texttt{LayerNorm}(\widetilde{\mathbf{Z}}_t^{(\ell)}+\texttt{FeedForward}(\widetilde{\mathbf{Z}}_t^{(\ell)})). 
\end{aligned}
\end{equation}

By computing the multiplication of $\sigma_1(\widetilde{\mathbf{K}}_e^{(\ell)})^{\mathsf{T}}$ and $\mathbf{Z}^{(\ell)}_t$ at first, the above process admits a $\mathcal{O}(ND_{\text{emb}})$ time complexity, which scales linearly with respect to the number of sensors. Since $\mathbf{E}$ is decoupled from temporal information, it is robust to missing values and reliable to infer a correlation map for global imputation. The full attention has the size $\mathbb{R}^{N\times (T\times D')}\times\mathbb{R}^{(T\times D')\times N}\rightarrow\mathbb{R}^{N\times N}$, while the embedded attention has $\mathbb{R}^{N\times D_{\text{emb}}}\times\mathbb{R}^{D_{\text{emb}}\times N}\rightarrow\mathbb{R}^{N\times N}$. In this sense, the attention map in Eq. \eqref{eq:graph_conv} act as a factorized low-rank approximation of full attention. We highlight this property by comparing the two formulations in the following analysis.

\begin{remark}[Difference between embedded attention and canonical self-attention]
Given a hidden state $\mathcal{Z}\in\mathbb{R}^{N\times T \times D'}$, the self-attention can be computed on the folded matrix $\mathbf{Z}\in\mathbb{R}^{N\times (TD')}$ as $\texttt{SelfAtten}(\mathbf{Z},\mathbf{Z},\mathbf{Z})=\sigma(\mathbf{ZW}_Q\mathbf{W}_K^{\mathsf{T}}\mathbf{Z}^{\mathsf{T}})\mathbf{Z}\mathbf{W}_V$. The resulting rank of the attention matrix obeys $r\leq\min\{N, TD'\}$. While the embedded attention has $\texttt{SelfAtten}(\mathbf{E},\mathbf{E},\mathbf{Z})=\sigma(\mathbf{E}\mathbf{W}_E)\sigma(\mathbf{E}\mathbf{W}_E)^{\mathsf{T}}\mathbf{Z}\mathbf{W}_V$. 
If we ignore the possible rank-increasing effect of \texttt{softmax}, the above calculation generates the output with rank $r\leq\min\{N, D_{\text{emb}}\}$. Since the dimension of the node embedding $D_{\text{emb}}$ is much smaller than the model dimension $D'$, the rank of the embedded attention map has a lower bound of rank than the full attention. In addition, the model still has a large feedforward dimension $D'$ to ensure capacity.
\end{remark}

\subsection{Fourier Imputation Loss}
As imputation model is typically optimized in a self-supervised manner, previous studies proposed adopting accumulated and hierarchical loss \cite{du2023saits,SPIN,GRIN} to improve the supervision of layerwise imputation. However, we argue that such designs are not necessary and may generate overfitting. Instead, we propose a novel Fourier imputation loss (FIL) combined with a simple supervision loss as task-specific biases to achieve effective training and generalization. 

\noindent\textbf{Self-Supervised Masked Learning}.
To create supervised samples, we randomly whiten a proportion of incomplete observations ($p_{\text{whiten}}$) during model training. This operation ensures the generalizability of the imputation model (see Section \ref{sec:random_mask}). We use a masking indicator $\mathbf{M}_{\text{whiten}}$ to denote these locations where the masked values are marked as ones and others as zeros.
Note that the supervision loss is only calculated on these manually whitened points, and models are forbidden to have access to the masked missing points used for evaluation. Therefore, the reconstruction loss of our model is a $\ell_1$ loss on the final imputation $\widehat{\mathbf{X}}$:
\begin{equation}
    \mathcal{L}_{\text{recon}}= \frac{1}{NT}\sum \Vert\mathbf{M}_{\text{whiten}}\odot(\widehat{\mathbf{X}}-\mathbf{Y})\Vert_1,
\end{equation}
where $\mathbf{Y}$ is the label used for training.

\noindent\textbf{Fourier Sparsity Regularization}.
As discussed above, we can obtain a reasonable imputation by constraining the rank of the estimated spatiotemporal tensor in the time domain. 
However, directly optimizing the rank of the tensor or matrix is challenging \cite{liu2012tensor}, as it includes some non-trivial or non-differentiable computations, such as truncated singular value decomposition (SVD). And the SVD of a $N\times T$ matrix costs $\mathcal{O}(\min\{N^2T,NT^2\})$ complexity, which can become a bottleneck when integrated with deep models. Fortunately, we can simplify this process using the following lemma.

\begin{lemma}[Equivalence between convolution nuclear norm and Fourier $\ell_1$ norm \cite{liu2022recovery,chen2022laplacian}] Given a smooth or periodic time series $\mathbf{x}\in\mathbb{R}^{T}$, its circulant (convolution) matrix $\mathcal{C}(\mathbf{x})\in\mathbb{R}^{T\times T}$ reflects the Tucker low-rankness, depicted by the convolutional nuclear norm. This property can be revealed by using the Discrete Fourier Transform (DFT). Let the DFT matrix be $\mathbf{U}\in\mathbb{C}^{T\times T}$, then the DFT is achieved by:
\begin{equation*}
\begin{aligned}
    \texttt{DFT}(\mathbf{x})=&\mathbf{U}\mathbf{x}=\mathbf{U}(\mathcal{C}(\mathbf{x})[:,0])
    = (\mathbf{U}\mathcal{C}(\mathbf{x}))[:,0].
\end{aligned}
\end{equation*}
As DFT diagonalizes the circulant matrix by $\mathcal{C}(\mathbf{x})=\mathbf{U}^{\mathsf{H}}\text{diag}(\sigma_1,\dots,\sigma_T)\mathbf{U}$ and $\mathbf{U}$ is a unitary matrix with the first column being ones, we have:
\begin{equation*}
\begin{aligned}
    (\mathbf{U}\mathcal{C}(\mathbf{x}))[:,0]&=(\text{diag}(\sigma_1,\sigma_2,\dots,\sigma_T)\mathbf{U})[:,0], \\
    &= [\sigma_1,\sigma_2,\dots,\sigma_T]^{\mathsf{T}}. \\
\end{aligned}
\end{equation*}
Therefore, we have $\Vert\texttt{DFT}(\mathbf{x})\Vert_0=\Vert[\sigma_1,\sigma_2,\dots,\sigma_T]^{\mathsf{T}}\Vert_0=\text{rank}(\mathcal{C}(\mathbf{x}))$, and $\Vert\texttt{DFT}(\mathbf{x})\Vert_1=\Vert\mathcal{C}(\mathbf{x})\Vert_*$.
\end{lemma}
This lemma means that we can efficiently obtain the singular values through DFT. The $\ell_0$ norm is exactly the matrix rank and the $\ell_1$ norm is equal to the nuclear norm $\Vert\mathcal{C}(\mathbf{x})\Vert_*$. 
Since $\Vert\mathcal{C}(\mathbf{x})\Vert_*$ serves as a convex surrogate of the rank of $\mathbf{x}$ in an approximate sense \cite{liu2022recovery}, we can equivalently achieve this goal by optimizing the Fourier $\ell_1$ norm. 
Considering the above equivalence, we can develop a sparsity-constrained loss function in the frequency domain:
\begin{equation}
\begin{aligned}
    \bar{\mathbf{X}} &= \mathbf{M}_{\text{missing}}\odot\widehat{\mathbf{X}}+(1-\mathbf{M}_{\text{missing}})\odot{\mathbf{Y}}, \\
    \mathcal{L}_{\text{FIL}}&=\frac{1}{NT}\sum\Vert\texttt{Flatten}(\texttt{FFT}(\bar{\mathbf{X}},\text{dim}=[0,1]))\Vert_1,\\
\end{aligned}
\end{equation}
where $\texttt{FFT}(\cdot)$ is the Fast Fourier Transform (FFT), $\texttt{Flatten}(\cdot):\mathbb{R}^{N\times T}\rightarrow\mathbb{R}^{NT}$ rearranges the tensor form and $\Vert\cdot\Vert_1$ is the vector $\ell_1$ norm. Since the spatiotemporal matrix can be regarded as a special RGB image from a global viewpoint, it also features a sparse Fourier spectrum in the space dimension \cite{liu2022recovery}. We apply the FFT on both the space and time axes and then flatten it into a long vector. $\mathcal{L}_{\text{FIL}}$ is in fact a unsupervised loss that encourages the imputed values to be naturally compatible with the observed values globally. 

Finally, the total loss function is formulated as:
\begin{equation}\label{eq:total_loss}
\mathcal{L}=\mathcal{L}_{\text{recon}}+\lambda\mathcal{L}_{\text{FIL}},
\end{equation}
where $\lambda$ is a weight hyperparameter. It is worth commenting that the two loss functions complement each other: $\mathcal{L}_{\text{recon}}$ prompts the model to reconstruct the masked observations as precisely as possible in the space-time domain and $\mathcal{L}_{\text{FIL}}$ generalizes on unobserved points with regularization on the spectrum. This makes \texttt{\acrshort{model}} work effectively in highly sparse observations.

\section{Empirical Evaluations}\label{experiments}
In this section, we evaluate our model on several well-known spatiotemporal benchmarks, comparing it with state-of-the-art baselines, and testing its generality on different scenarios. Then comprehensive analysis and case studies are provided. A brief summary of the adopted datasets is shown in Tab. \ref{tab_volume_datasets}. Detailed descriptions of experimental settings are provided in Section \ref{Appendix:settings}. PyTorch implementations are available at \underline{\url{https://github.com/tongnie/ImputeFormer}}.

\begin{table}[!htb]
\renewcommand{\arraystretch}{0.9} 
\setlength{\abovecaptionskip}{0.cm}
\setlength{\belowcaptionskip}{-0.0cm}
\caption{Statistics of benchmark datasets.}
\label{tab_volume_datasets}
\centering
\setlength{\tabcolsep}{10pt}
\resizebox{1\columnwidth}{!}{
\begin{tabular}{l|c c c c c}
\toprule
 Datasets & Type & Steps & Nodes & Interval \\
\toprule
\texttt{METR-LA} &  Traffic speed & 34,272 & 207 & 5 min  \\
\texttt{PEMS-BAY} & Traffic speed & 52,128 & 325 & 5 min  \\
\texttt{PEMS03} &  Traffic volume & 26,208 & 358 & 5 min  \\
\texttt{PEMS04} &  Traffic volume & 16,992 & 307 & 5 min \\
\texttt{PEMS07} &  Traffic volume & 28,224 & 883 & 5 min  \\
\texttt{PEMS08} &  Traffic volume & 17,856 & 170 & 5 min  \\
\midrule
\texttt{SOLAR} &  Power production & 52,560 & 137 & 10 min  \\
\texttt{CER-EN} &  Energy consumption & 8,868 & 435 & 30 min  \\
\midrule
\texttt{AQI} &  Air pollutant & 8,760 & 437 & 60 min  \\
\texttt{AQI36} &  Air pollutant & 8,760 & 36 & 60 min  \\
\bottomrule
\end{tabular}}
\end{table}

\subsection{Results on Traffic Benchmarks}\label{results:traffic}
The imputation results on traffic speed and volume data are given in Tab. \ref{tab:taffic_results}. As can be seen, \texttt{\acrshort{model}} consistently achieves the best performance in all traffic benchmarks. Two strong competitors \texttt{GRIN} and \texttt{SPIN} show promising results in traffic speed data, which align with the results of their respective papers \cite{GRIN,SPIN}. However, their performance is inferior on volume datasets and is surpassed by simple baselines such as \texttt{ST-Transformer} and \texttt{Bi-MPGRU}. Compared to deep models, pure low-rank methods, such as matrix factorization and tensor completion, are less effective due to limited capacity.
As for missing patterns, the structured block missing is more challenging than the point missing pattern. For instance, the vanilla Transformer is competitive in the point missing case, while it is ineffective in block missing case. Generally, \texttt{\acrshort{model}} outperforms others by a large margin in this tricky scenario.

\begin{table*}[!htbp]
\renewcommand{\arraystretch}{0.9} 
\setlength{\abovecaptionskip}{0.cm}
\setlength{\belowcaptionskip}{-0.0cm}
\caption{{Results (in terms of MAE) on \texttt{METR-LA}, \texttt{PEMS-BAY}, \texttt{PEMS03}, \texttt{PEMS04}, \texttt{PEMS07} and \texttt{PEMS08} traffic benchmarks.}}
\begin{center}
\setlength{\tabcolsep}{5pt}
\resizebox{1\textwidth}{!}{
\begin{tabular}{l|c|c|c|c|c|c|c|c|c|c|c|c}
\toprule
 & \multicolumn{6}{c|}{Point missing} & \multicolumn{6}{c}{Block missing}  \\
\hline
\multicolumn{1}{l|}{Models} & \multicolumn{1}{c|}{\texttt{PEMS-BAY}} & \multicolumn{1}{c|}{\texttt{METR-LA}} & \multicolumn{1}{c|}{\texttt{PEMS03}} & \multicolumn{1}{c|}{\texttt{PEMS04}} & \multicolumn{1}{c|}{\texttt{PEMS07}} & \multicolumn{1}{c|}{\texttt{PEMS08}} & \multicolumn{1}{c|}{\texttt{PEMS-BAY}} & \multicolumn{1}{c|}{\texttt{METR-LA}} & \multicolumn{1}{c|}{\texttt{PEMS03}} & \multicolumn{1}{c|}{\texttt{PEMS04}}  & \multicolumn{1}{c|}{\texttt{PEMS07}} & \multicolumn{1}{c}{\texttt{PEMS08}}\\
\hline
{Average} & 5.45  & 7.52   & 85.30  &  103.61 & 122.35 &  89.51 & 5.48 & 7.43 & 85.56  & 103.82
 &123.05 &89.42\\
 {MICE \cite{van2011mice}} & 2.82  & 2.89  &  20.07 & 28.60 & 37.11 & 30.26  & 2.36 &  2.73 & 21.90  & 32.45 & 37.20 & 26.66 \\
 {TRMF \cite{yu2016temporal}} & 2.10  & 3.51   & 18.80  & 24.34  & 29.06 &  20.27 & 2.09 &  3.36 & 18.71  & 24.47 &29.42  &19.80 \\
{LRTC-AR \cite{chen2021low}} & 0.94  & 2.14 & 15.52 & 22.11 & 27.60 & 19.33 & 4.05 &  5.35 & 17.59 & 24.08 & 27.82 & 19.95\\
{Bi-MPGRU} & 0.72  & 2.00  & 11.23  & \color{brown}\underline{15.84}  & 15.66 &  \color{brown}\underline{11.90}   & 1.41 &  2.33 & 13.87  & \color{brown}\underline{19.81} &21.12 & \color{brown}\underline{15.89} \\
{rGAIN \cite{yoon2018gain}} & 1.90  & 2.81 & 13.32  & 22.86 & 24.41 &  16.33  & 2.21 &  2.95  & 14.85  & 23.26 & 26.69 & 27.12\\
{BRITS \cite{cao2018brits}}  & 1.84  & 2.42   & 12.74  & 20.00  & 23.97  & 15.78  & 1.91  &  2.40 & 12.93  & 19.80 & 23.26 &  16.37 \\
{SAITS \cite{du2023saits}} & 1.33 & 2.25  & 12.40 & 20.23 & 22.81 &  15.12 & 1.58 &  2.32 & 12.43  & 20.35 & 22.82 & 16.80\\
{Transformer \cite{transformer}} & 0.76  & 2.18   & 12.04  & 16.76  & 16.86  &  12.58 & 1.69  &  3.58 & 24.07  & 29.63 & 33.14 & 25.61 \\
{ST-Transformer} & 0.75  &  2.19   &  11.44 & 16.22 & 15.84  &  12.10 & 1.71  &  3.58  & 23.55  &29.17  & 32.14 & 24.67 \\
{TIDER \cite{liu2022multivariate}} & 1.43  &  2.68 & 15.02&22.17 & 21.38& 18.46& 2.46  &  4.95 &21.12 & 23.74 & 28.66&21.00\\
{TimesNet \cite{wu2022timesnet}} & 1.47  &  2.93   & 14.99 & 20.40  & 22.00  & 16.53  & 2.73  &  4.79  & 44.85 & 51.05 &60.90  & 45.78\\
{GRIN \cite{GRIN}} &  \color{brown}\underline{{0.68}}  & \color{brown}\underline{{1.91}}   &  \color{brown}\underline{10.31}  & 16.25 & \color{brown}\underline{11.90}  &  12.33  & 1.20  &  2.08 & \color{brown}\underline{12.28} & 23.23 & \color{brown}\underline{16.04} &19.69 \\
{SPIN \cite{SPIN}}  & 0.79  &  1.93   & 12.85  &  18.96  & 17.61 &  15.02 & \color{brown}\underline{1.13} &  \color{brown}\underline{2.02}  & 14.68 & 19.85 & 16.99 & 16.81 \\
\hline
\multirow{2}{*}{\textbf{\gls{model}}} & \color{purple}\textbf{0.64} & \color{purple}\textbf{1.80} & \color{purple}\textbf{8.23} & \color{purple}\textbf{14.92} & \color{purple}\textbf{11.38} & \color{purple}\textbf{11.01} & \color{purple}\textbf{0.95} & \color{purple}\textbf{1.86} & \color{purple}\textbf{9.02} & \color{purple}\textbf{16.83} & \color{purple}\textbf{13.82} & \color{purple}\textbf{12.50}\\
& \cellcolor{green!30}$\textbf{5.9\%}\downarrow$ & \cellcolor{green!30}{$\textbf{5.8\%}\downarrow$} & \cellcolor{green!30}{$\textbf{20.2\%}\downarrow$} &  \cellcolor{green!30}{$\textbf{5.8\%}\downarrow$} &  \cellcolor{green!30}{$\textbf{4.4\%}\downarrow$} &  \cellcolor{green!30}{$\textbf{7.5\%}\downarrow$} &  
\cellcolor{green!30}{$\textbf{15.9\%}\downarrow$} & \cellcolor{green!30}{$\textbf{7.9\%}\downarrow$}& \cellcolor{green!30}{$\textbf{26.5\%}\downarrow$} &  \cellcolor{green!30}{$\textbf{15.0\%}\downarrow$} &  \cellcolor{green!30}{$\textbf{13.8\%}\downarrow$} &  \cellcolor{green!30}{$\textbf{21.3\%}\downarrow$} \\
\bottomrule
\end{tabular}}
\end{center}
\label{tab:taffic_results}
\end{table*}

\subsection{Results on Environmental and Energy Data}
By exploiting the underlying low-rank structures, \texttt{\acrshort{model}} can serve as a general imputer in a variety of spatiotemporal data. 
To demonstrate its versatility, we perform experiments on other spatiotemporal data, including energy and environmental data.  
Results are given in Tab. \ref{tab:expotherresults}. It is observed that \texttt{\acrshort{model}} exhibits superiority in other spatiotemporal datasets beyond traffic data. In particular, the correlation of solar stations cannot be described by physical distance and can be inferred from the data. After comparing the performance of \texttt{SAITS}, \texttt{Transformer}, \texttt{ST-Transformer}, and \texttt{\acrshort{model}}, it can be concluded that direct attention computations on both temporal and spatial dimensions are less beneficial than the low-rank attention. Furthermore, the spatial correlation of energy production is less pronounced. Canonical attention on the spatial axis can be redundant and generate spurious correlations. The use of embedded attention in our model can alleviate this issue.

\begin{table}[!htbp]
\renewcommand{\arraystretch}{0.9} 
\setlength{\abovecaptionskip}{0.cm}
\setlength{\belowcaptionskip}{-0.0cm}
\caption{{Results (in terms of MAE) on \texttt{AQI}, \texttt{Solar}, and \texttt{CER-EN} benchmarks. For \texttt{Solar} data, we compare the performances of baselines that are independent of the predefined graphs.}}
\begin{center}
\setlength{\tabcolsep}{4pt}
\resizebox{1\columnwidth}{!}{
\begin{tabular}{l|c|c|c|c|c|c}
\toprule
 & \multicolumn{2}{c|}{\texttt{SOLAR}} & \multicolumn{2}{c|}{\texttt{CER-EN}} & \multicolumn{2}{c}{Simulated faults}  \\
\hline
\multicolumn{1}{l|}{Models} & \multicolumn{1}{c|}{\scalebox{0.9}{\makecell{Point \\ missing}}} & \multicolumn{1}{c|}{\scalebox{0.9}{\makecell{Block\\ missing}}} & \multicolumn{1}{c|}{\scalebox{0.9}{\makecell{Point \\ missing}}} & \multicolumn{1}{c|}{\scalebox{0.9}{\makecell{Block\\ missing}}} & \multicolumn{1}{c|}{{\texttt{AQI36}}} & \multicolumn{1}{c}{{\texttt{AQI}}}  \\
\hline
{Average}  & 7.60 & 7.56 & 0.583 & 0.596 & 61.81  & 43.78 \\
{MICE }  &1.59  & 1.58 & 0.535 & 0.555 & 38.90  & 29.12 \\
{TRMF}  & 2.44 & 2.35 & 0.557 & 0.559 & 41.91  & 27.67 \\
{Bi-MPGRU}  & N.A. & N.A. & 0.247 & 0.349 &  12.02  & 15.41 \\
{rGAIN} & 1.52 & 1.64 & 0.418 & 0.440 & 15.69  & 22.13 \\
{BRITS} & 1.28 & 1.34 & 0.351 & 0.366 & 14.74  & 20.72    \\
{SAITS}  & \color{brown}\underline{0.98} & \color{brown}\underline{1.25} & 0.341 & 0.368 & 19.79  & 21.09 \\
{Transformer}  &  2.19 & 3.58 & 0.254 & 0.353 & 14.99  & 17.04  \\
{ST-Transformer} &2.17  &  3.57 & 0.251 & 0.351 & 13.27  & 18.55  \\
{TIDER}   & 2.84 & 3.87 & 0.336 & 0.377 & 32.85 & 18.11  \\
{TimesNet} & 2.93 & 4.73  &0.328 & 0.460 & 32.30 & 28.99  \\
{GRIN} &N.A.  & N.A. & \color{purple}\textbf{0.235} & \color{brown}\underline{0.341} & 12.08 & 14.51  \\
{SPIN} & N.A. &  N.A. &  OOM & OOM & \color{brown}\underline{11.89} & \color{brown}\underline{14.31}  \\
\hline
\multirow{2}{*}{\textbf{\gls{model}}} & \color{purple}\textbf{0.51} & \color{purple}\textbf{0.89} & \color{brown}\underline{0.236} & \color{purple}\textbf{0.296} & \color{purple}\textbf{11.58} & \color{purple}\textbf{13.40} \\
&  \cellcolor{green!30}{$\textbf{48.0\%}\downarrow$}&  \cellcolor{green!30}{$\textbf{28.8\%}\downarrow$}&  \cellcolor{red!20}{$\textbf{0.4\%}\uparrow$}&  \cellcolor{green!30}{$\textbf{13.2\%}\downarrow$}&  \cellcolor{green!30}{$\textbf{2.6\%}\downarrow$}&  \cellcolor{green!30}{$\textbf{6.4\%}\downarrow$}\\
\bottomrule
\end{tabular}}
\label{tab:expotherresults}
\end{center}
\end{table}

\subsection{Ablation Study}
To justify the rationale of model designs, we conduct ablation studies on the model structure. Results are shown in Tab. \ref{tab:ablation}. Several intriguing findings can be observed: (1) After removing any of the temporal and spatial attention modules, the performance degenerates substantially; especially, the \textit{spatial interaction contributes to the inference of block missing patterns significantly}, while the \textit{temporal modules are crucial for point missing scenarios}. (2) \textit{The incorporation of \texttt{MLP} benefits little for the imputation}, which validates our argument in Section \ref{appendix: input_embedding}. (3) Compared to hierarchical loss on supervised points, FIL generalizes on the unobserved points and effectively reduces the estimation errors. 

\begin{table}[!htbp]
\renewcommand{\arraystretch}{0.9} 
\setlength{\abovecaptionskip}{0.cm}
\setlength{\belowcaptionskip}{-0.0cm}
\caption{Ablations studies on \texttt{\acrshort{model}}.}
\label{tab:ablation}
\centering
\setlength{\tabcolsep}{4pt}
\resizebox{1\columnwidth}{!}{
    \begin{tabular}{c|c|c|cc|cc}
    \toprule
    \multirow{2}{*}{Variation} &\multicolumn{2}{c|}{Component} & \multicolumn{2}{c|}{\texttt{PEMS08}} & \multicolumn{2}{c}{\texttt{METR-LA}}\\
    \cmidrule(lr){2-3} \cmidrule(lr){4-5} \cmidrule(lr){6-7} 
    & Spatial& Temporal& Point & Block & Point & Block   \\
    \midrule
     \textbf{\acrshort{model}} & \cellcolor{green!30}\textbf{Attention} & \cellcolor{green!30}\textbf{Attention} & \cellcolor{green!30}\textbf{11.24} & \cellcolor{green!30}\textbf{12.86} & \cellcolor{green!30}\textbf{1.80} & \cellcolor{green!30}\textbf{1.88} \\
     \midrule
     \multirow{3}{*}{Replace}& Attention & MLP & 16.95 & 17.11 & 2.39 & 2.28 \\
     & MLP & Attention & 12.84 & 17.42  & 2.20 & 2.92  \\
     & MLP & MLP & 34.72 & 34.41 & 5.80 & 5.79 \\
    \midrule
    \multirow{2}{*}{w/o} & Attention & w/o & 17.06 & 17.13 & 2.39 & 2.28 \\
     & w/o & Attention & 12.87 &  17.44 & 2.21 &  2.93 \\
     \cmidrule[1.pt]{1-7}
    \multirow{2}{*}{Loss function} & \multicolumn{2}{c|}{w/o FIL} & 11.63 & 13.35 & 1.85  & 1.93 \\
     & \multicolumn{2}{c|}{ Hierarchical loss} & 11.35 & 13.07 & 1.84 & 1.92  \\
     \cmidrule[1.pt]{1-7}
    \multirow{3}{*}{Architecture} & \multirow{2}{*}{Order} & \multicolumn{1}{c|}{T-S} & 11.26 & 13.16 & 1.80  & 1.87 \\
     & & \multicolumn{1}{c|}{S-T} & 11.30 & 13.13 & 1.80 & 1.88  \\
     \cmidrule{2-7}
    & \multicolumn{2}{c|}{Joint ST} &  17.50 & 20.00 & 1.94 & 2.58  \\
    \bottomrule
    \end{tabular}}
\end{table}

\begin{figure}[!htbp]
	\centering
		\includegraphics[scale=0.28]{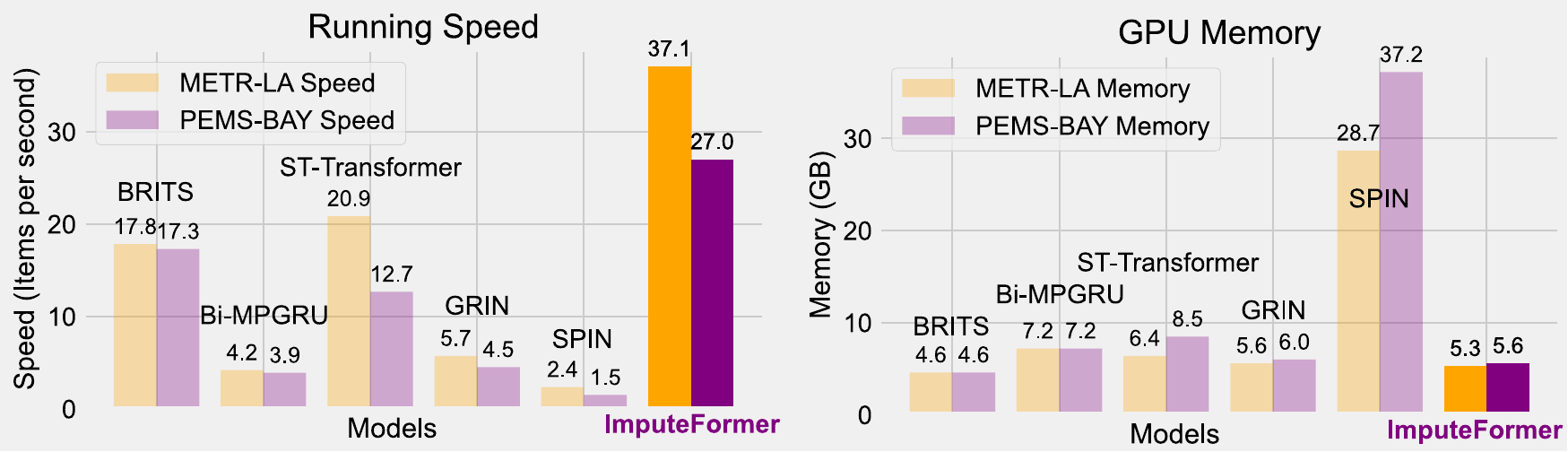}
  \captionsetup{skip=1pt}
	\caption{Comparison of computational efficiency.}
	\label{FIG:efficiency}
\end{figure}

\subsection{Model Efficiency}
We evaluate the computational efficiency of different architectures in Fig. \ref{FIG:efficiency}.
Intuitively, \texttt{\acrshort{model}} exhibits high training efficiency. Due to the low-rank design philosophy, \texttt{\acrshort{model}} is approximately 15 times faster than the state-of-the-art Transformer baseline (\texttt{SPIN}).
It is also cost-effective in GPU memory consumption.

\subsection{Robustness and Versatility Analysis}
\noindent\textbf{{Inference under Different Missing Rates}}. \label{sec:masking_rate}
Deep imputation models are subject to the distribution shift problem between training and testing datasets. A desirable characteristic is that a model can deal with different missing patterns during inference. Therefore, we consider a challenging scenario in which a model is trained with a fixed missing rate but evaluated on different scenarios with varying missing rates. This constructs a zero-shot transfer evaluation. It is noteworthy in Tab. \ref{tab:varying_missing} that both \texttt{\acrshort{model}} and \texttt{SPIN} are more robust than other baselines in these scenarios. 
RNNs and vanilla Transformers can overfit the training data with a fixed data missing pattern, thereby showing inferior generalization ability.

\begin{table}[!htbp]
\renewcommand{\arraystretch}{0.9} 
\setlength{\abovecaptionskip}{0.cm}
\setlength{\belowcaptionskip}{-0.0cm}
\caption{Inference under varying missing rate with a single trained model (Zero-shot).}
\centering
\setlength{\tabcolsep}{8pt}
\resizebox{1\columnwidth}{!}{
\begin{tabular}{l|c|c|c|c|c|c}
\toprule
 & \multicolumn{3}{c|}{\texttt{PEMS08}} & \multicolumn{3}{c}{\texttt{METR-LA}}  \\
\hline
\multirow{2}{*}{Models} & \multicolumn{3}{c|}{Missing rate} & \multicolumn{3}{c}{Missing rate}  \\
\cmidrule(lr){2-4} \cmidrule(lr){5-7}
& \multicolumn{1}{c|}{$50\%$} & \multicolumn{1}{c|}{\texttt{$75\%$}} & \multicolumn{1}{c|}{$95\%$} & \multicolumn{1}{c|}{$50\%$} & \multicolumn{1}{c|}{\texttt{$75\%$}} & \multicolumn{1}{c}{$95\%$}\\
\hline
{BRITS } & 17.21 & 22.01  & 52.78 & 2.61 & 3.04  & 5.11   \\
{SAITS}  & 16.03 & 31.32 &83.79  & 2.44 & 3.37  & 6.80  \\
{ST-Transformer} & 11.65  & 13.11 & 39.95 & 2.32 & 2.72  &5.16    \\
{GRIN} &13.25   & 16.06 &42.61 & 2.06 & 2.39  &  4.07  \\
{SPIN } & 15.13  & 15.51 & 18.30 & 2.11 & 2.34  & 3.03  \\
\hline
\textbf{\gls{model}} &\cellcolor{green!30}\textbf{11.52} & \cellcolor{green!30}\textbf{12.18} & \cellcolor{green!30}\textbf{17.35} & \cellcolor{green!30}\textbf{1.96} & \cellcolor{green!30}\textbf{2.17} & \cellcolor{green!30}\textbf{2.79} \\
\bottomrule
\end{tabular}}
\label{tab:varying_missing}
\end{table}

\begin{figure}[!htbp]
\centering
\begin{subfigure}[b]{0.49\linewidth}
\centering
\includegraphics[scale=0.27]{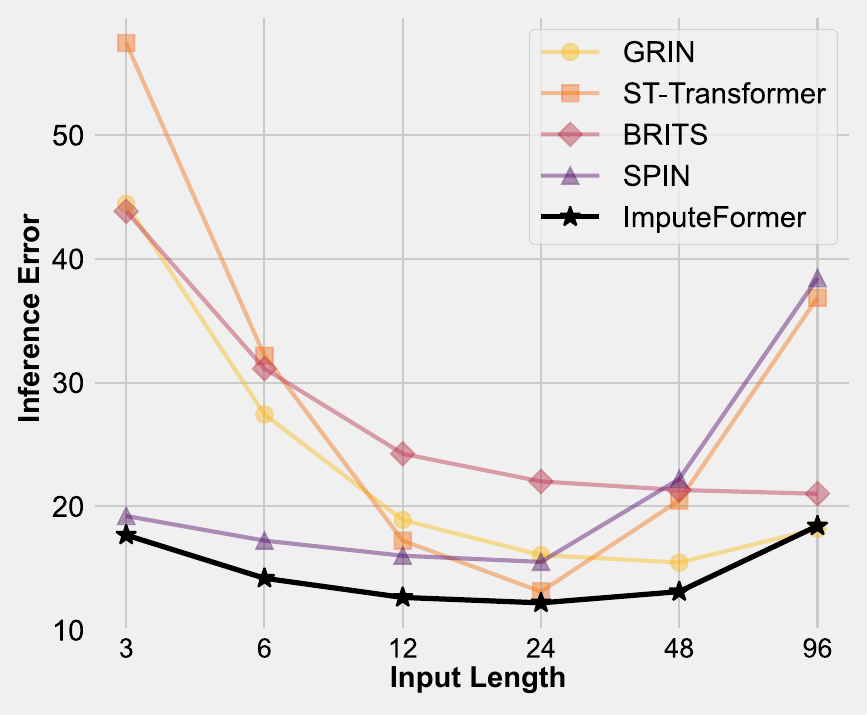}
\captionsetup{skip=1pt}
\caption{PEMS08}
\end{subfigure}
\hfill
\begin{subfigure}[b]{0.49\linewidth}  
\centering
\includegraphics[scale=0.27]{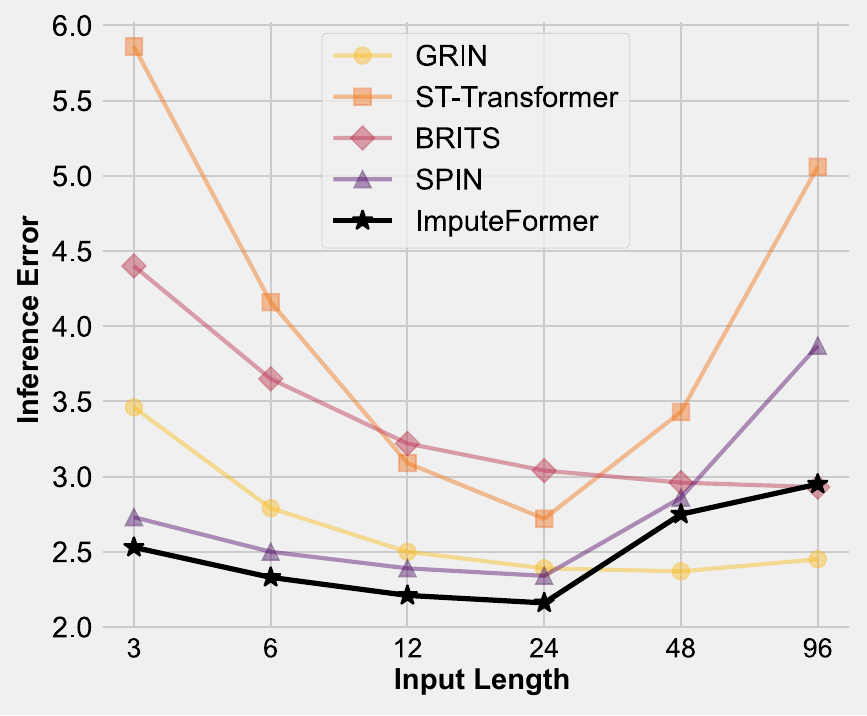}
\captionsetup{skip=1pt}
\caption{METR-LA}
\end{subfigure}
\captionsetup{skip=1pt}
\caption{Inference under different lengths of input sequence with a single trained model (zero-shot).}
\label{fig:varying_length}
\end{figure}

\noindent\textbf{Inference with Varying Sequence Length}.
In reality, the imputation model can face time series with different lengths and sampling frequencies. We can adopt a well-trained model to perform inference on varying sequence length. 
Results are shown in Fig. \ref{fig:varying_length}. It is obvious that \texttt{\acrshort{model}} can readily generalize to sequences with different lengths and more robust than other models.

\noindent\textbf{Dealing with Highly Sparse Observation}.
To evaluate the performance on highly sparse data, we further train and test models with lower observation rates. Results are shown in Tab. \ref{tab:sparse_result}. Generally speaking, both Transformer- and RNN-based models are susceptible to sparse training data. Due to the low-rank constraints on the attention matrix and loss function, our model is more robust with highly sparse data. Since the attention map in \texttt{SPIN} is calculated only at the observed points, it is more stable than other baselines. But our model consistently achieves lower imputation errors.

\begin{table}[!htbp]
\renewcommand{\arraystretch}{0.9} 
\setlength{\abovecaptionskip}{0.cm}
\setlength{\belowcaptionskip}{-0.0cm}
\caption{{Results on \texttt{PEMS08} data with sparse observations (Training from scratch).}}
\centering
\setlength{\tabcolsep}{12pt}
\resizebox{1\columnwidth}{!}{
\begin{tabular}{l|c|c|c|c}
\toprule
 & \multicolumn{4}{c}{Missing rate}   \\
\hline
\multicolumn{1}{l|}{Models} & \multicolumn{1}{c|}{$60\%$} & \multicolumn{1}{c|}{$70\%$} & \multicolumn{1}{c|}{\texttt{$80\%$}} & \multicolumn{1}{c}{$90\%$} \\
\hline
{BRITS} & 18.60 & 19.75  & 21.44  & 24.17    \\
{SAITS} &  17.53 & 18.24  & 19.39  & 21.27    \\
{ST-Transformer}& 13.67 & 14.32  & 15.86  & 23.98 \\
{GRIN} & 14.04 & 15.01  & 17.26 & 25.47 \\
{SPIN } & 13.64 & 14.30 & 15.19   & 17.13 \\
\hline
\textbf{\gls{model}} & \cellcolor{green!30}\textbf{12.57} & \cellcolor{green!30}\textbf{13.17} & \cellcolor{green!30}\textbf{13.98} & \cellcolor{green!30}\textbf{15.94} \\
\bottomrule
\end{tabular}}
\label{tab:sparse_result}
\end{table}

\begin{table}[!htbp]
\renewcommand{\arraystretch}{0.9} 
\setlength{\abovecaptionskip}{0.cm}
\setlength{\belowcaptionskip}{-0.0cm}
\caption{Results on \texttt{PEMS03} with various masking strategies.}
\centering
\setlength{\tabcolsep}{12pt}
\resizebox{1\columnwidth}{!}{
\begin{tabular}{l|c|c|c|c}
\toprule
 Point missing & \multicolumn{4}{c}{Masking Probability}   \\
\hline
\multicolumn{1}{l|}{Models} & \multicolumn{1}{c|}{$25\%$} & \multicolumn{1}{c|}{\texttt{$50\%$}} & \multicolumn{1}{c|}{$75\%$} & \multicolumn{1}{c}{Combined} \\
\hline
{Bi-MPGRU}  & 11.30 & 11.52  & 12.20  & 11.48 \\
{BRITS } & 13.06 & 13.86  & 16.06  & 13.70  \\
{SAITS}  & 12.42 & 16.13 & 21.63  & 12.61 \\
{ST-Transformer} & 11.19  & 11.43 & 12.22  & 11.39  \\
{GRIN} & 9.55  & 9.74 & 10.39 & 9.72  \\
{SPIN } & 11.08 & 11.21   & 13.85 & 12.04  \\
\hline
\textbf{\gls{model}} & \cellcolor{green!30}\textbf{7.66} & \cellcolor{green!30}\textbf{8.16} & \cellcolor{green!30}\textbf{11.44} & \cellcolor{green!30}\textbf{8.45} \\
\hline
\end{tabular}}
\centering
\setlength{\tabcolsep}{12pt}
\resizebox{1\columnwidth}{!}{
\begin{tabular}{l|c|c|c|c}
\toprule
 Block missing& \multicolumn{4}{c}{Masking Probability}   \\
\hline
\multicolumn{1}{l|}{Models} & \multicolumn{1}{c|}{$25\%$} & \multicolumn{1}{c|}{\texttt{$50\%$}} & \multicolumn{1}{c|}{$75\%$} & \multicolumn{1}{c}{Combined} \\
\hline
{Bi-MPGRU}  & 13.32 & 13.36  & 13.96  & 13.33 \\
{BRITS } & 12.26 & 13.01  & 15.58  & 12.63    \\
{SAITS}  & 12.35 & 15.73 & 20.14  & 12.32 \\
{ST-Transformer} &23.51  &  23.76 & 23.90  & 23.26  \\
{GRIN} &11.94  & 12.05 & 12.68 & 11.99  \\
{SPIN } & 13.10 & 13.68   & 13.84 & 13.97  \\
\hline
\textbf{\gls{model}} & \cellcolor{green!30}\textbf{8.89} & \cellcolor{green!30}\textbf{9.23} & \cellcolor{green!30}\textbf{16.96} & \cellcolor{green!30}\textbf{8.80} \\
\bottomrule
\end{tabular}}
\label{tab:whiten_prob}
\end{table}

\noindent\textbf{Random Masking in Training}. \label{sec:random_mask}
Random masking strategy is used to create supervised samples for model training. Therefore, the distribution of masking samples of training data and missing observations of testing data should be close to ensure good performance \cite{nie2023towards}. However, it can be difficult to know exactly the missing patterns or missing rates in advance in many scenarios. Therefore, a proper masking strategy is of vital importance. To evaluate the impact of the masking rate in training data, we further consider four different masking strategies during model training: the masking rates are set to $0.25, 0.5, 0.75$, and a combination of them $[0.25, 0.5, 0.75]$ respectively. 
As shown in Tab. \ref{tab:whiten_prob}, most models perform best when the masking rate is close to the missing rate in the point missing scenario (e.g., $25\%$). However, when the missing rate is unclear due to randomness within the failure generation process, such as the block missing, the combination strategy is more advantageous. For example, Transformers including \texttt{ST-Transformer}, \texttt{SAITS}, and \texttt{\acrshort{model}} can benefit from this strategy. More importantly, such a hybrid masking method enables the model to work successfully on varying observation rates during inference.

\begin{figure}[!htbp]
\centering
\begin{subfigure}[b]{0.45\linewidth}
\centering
\includegraphics[scale=0.25]{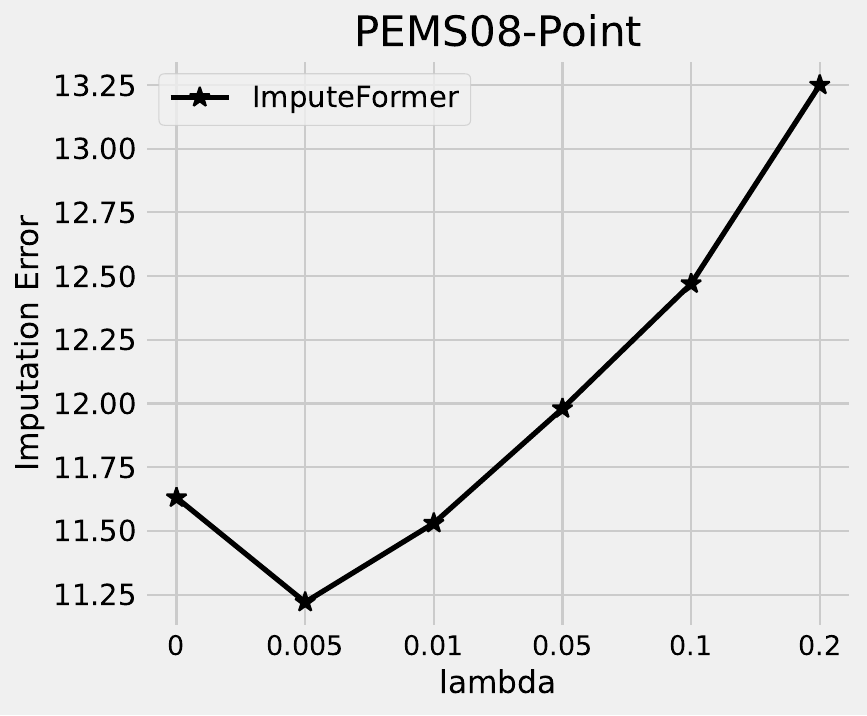}
\captionsetup{skip=1pt}
\caption{PEMS08, point missing}
\end{subfigure}
\begin{subfigure}[b]{0.45\linewidth}  
\centering
\includegraphics[scale=0.25]{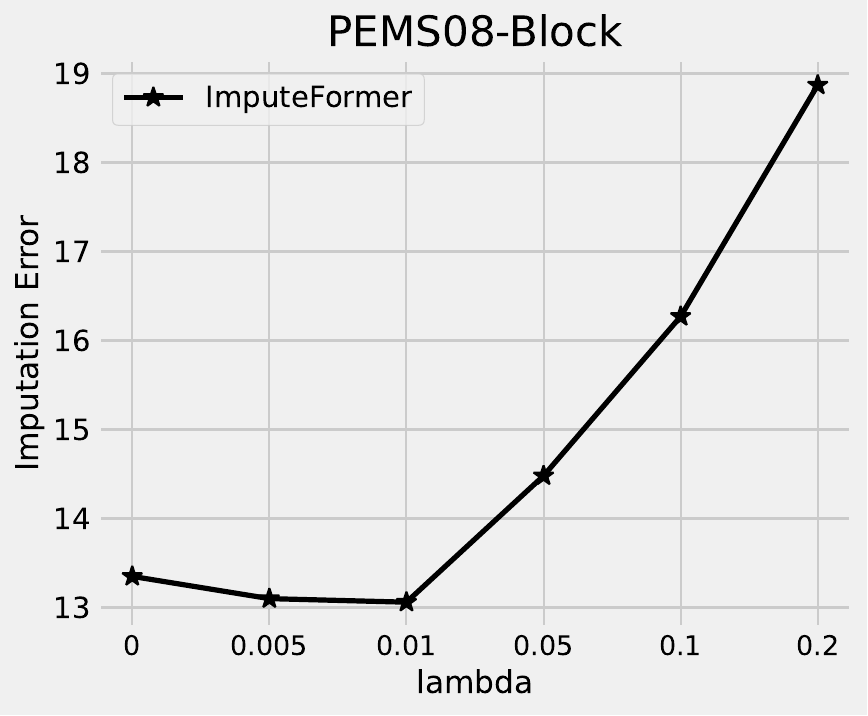}
\captionsetup{skip=1pt}
\caption{PEMS08, block missing}
\end{subfigure}
\captionsetup{skip=1pt}
\caption{Impact of $\lambda$ in FIL.}
\label{fig:lambda}
\end{figure}

\noindent\textbf{Impact of Fourier Imputation Loss}. \label{Appendix: FIL}
We study the impact of hyperparameter $\lambda$ in Eq. \eqref{eq:total_loss}. Fig. \ref{fig:lambda} displays the imputation error under different $\lambda$ values. The imputation model can benefit from the FIL design, and a too large penalty on the sparsity of the spectrum can lead to a smooth and inaccurate reconstruction.

\subsection{Case Studies: Interpretability}
This section studies the interpretability using data examples.

\begin{figure}[!htbp]
\centering
\begin{minipage}[c]{0.49\columnwidth}
  \centering
  \includegraphics[scale=0.24]{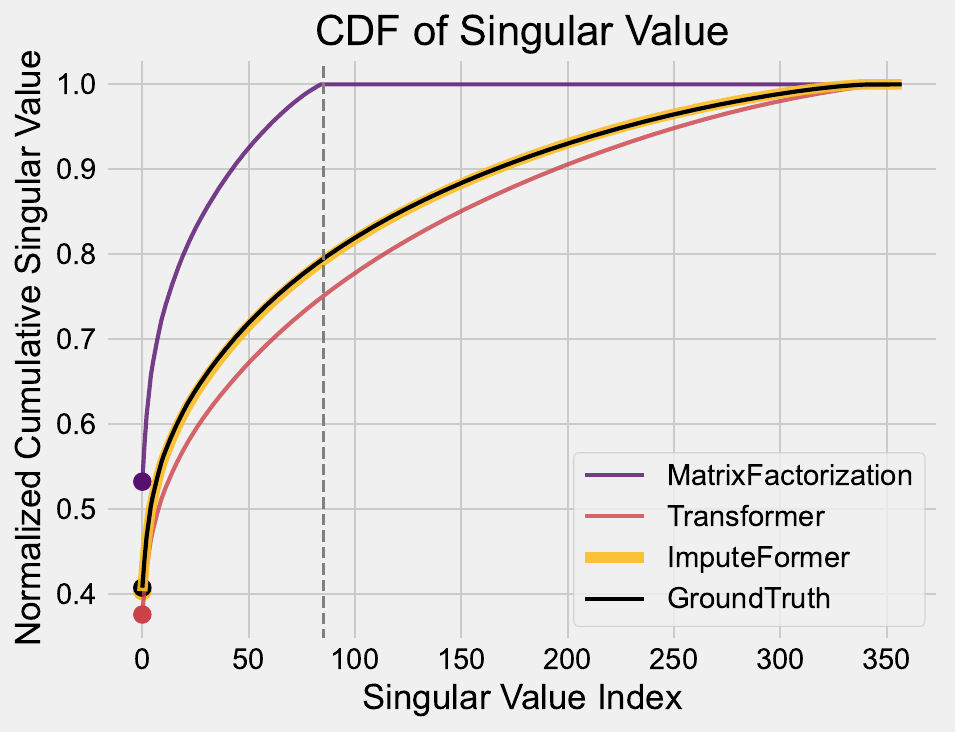}
  \captionsetup{skip=1pt}
  \caption{Cumulative distribution of singular values.}
  \label{fig:svd_cdf}
\end{minipage}\hfill
\begin{minipage}[c]{0.49\columnwidth}
  \centering
  \includegraphics[scale=0.23]{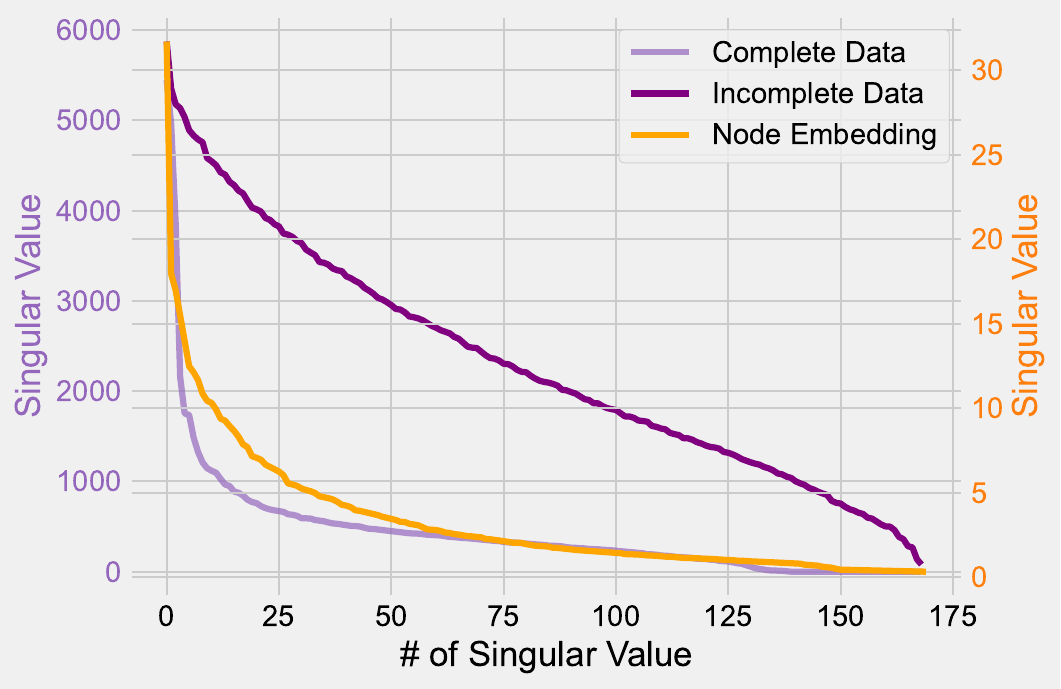}
  \captionsetup{skip=1pt}
  \caption{Singular spectrum of data and node embedding.}
  \label{fig:node_embed}
\end{minipage}
\end{figure}

\noindent\textbf{Spectrum Analysis}.
To corroborate the hypothesis that our model has the merits of both deep learning and low-rank methods, we analyze the singular value (SV) spectrum of the imputations. Fig. \ref{fig:svd_cdf} shows the cumulative SV distribution of different competing models. \texttt{\acrshort{model}} has a close SV cumulative distribution to complete data, and the first 85 SVs can account for $80\%$ of the energy.
There exist two additional interesting observations: (1) deep learning models without explicit low-rank modeling such as canonical Transformers downplay the role of the first few dominant SVs; (2) pure low-rank models such as MF generate an oversmoothing result that too much energy is constrained to the first part of spectrum. Thus, we can ascribe the desirable performance of our model to the \textit{good balance of significant signals and high-frequency noise}.

\begin{figure}[!htbp]
\centering
\includegraphics[scale=0.38]{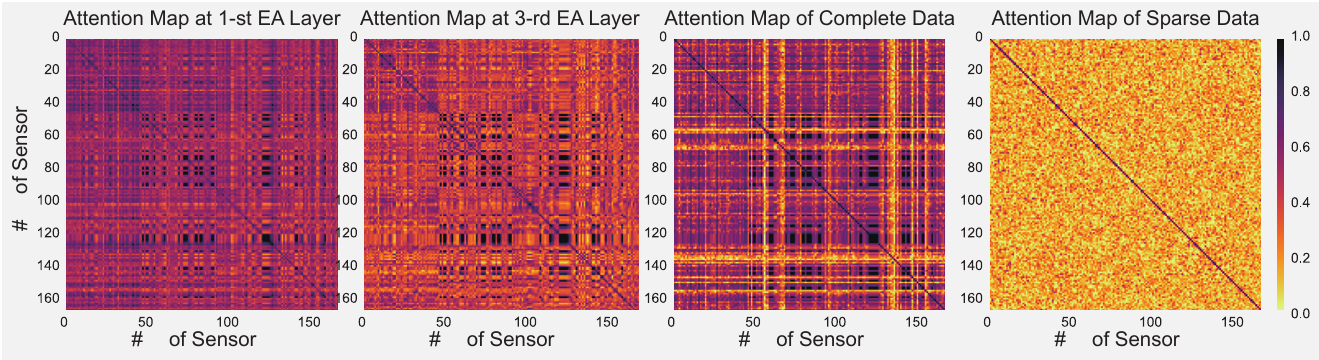}
\captionsetup{skip=1pt}
\caption{Multivariate attention maps of \texttt{PEMS08} data.} 
\label{fig:sptial_att}
\end{figure}

\begin{figure}[!htbp]
\centering
\begin{minipage}[c]{0.49\columnwidth}
  \centering
  \includegraphics[scale=0.2]{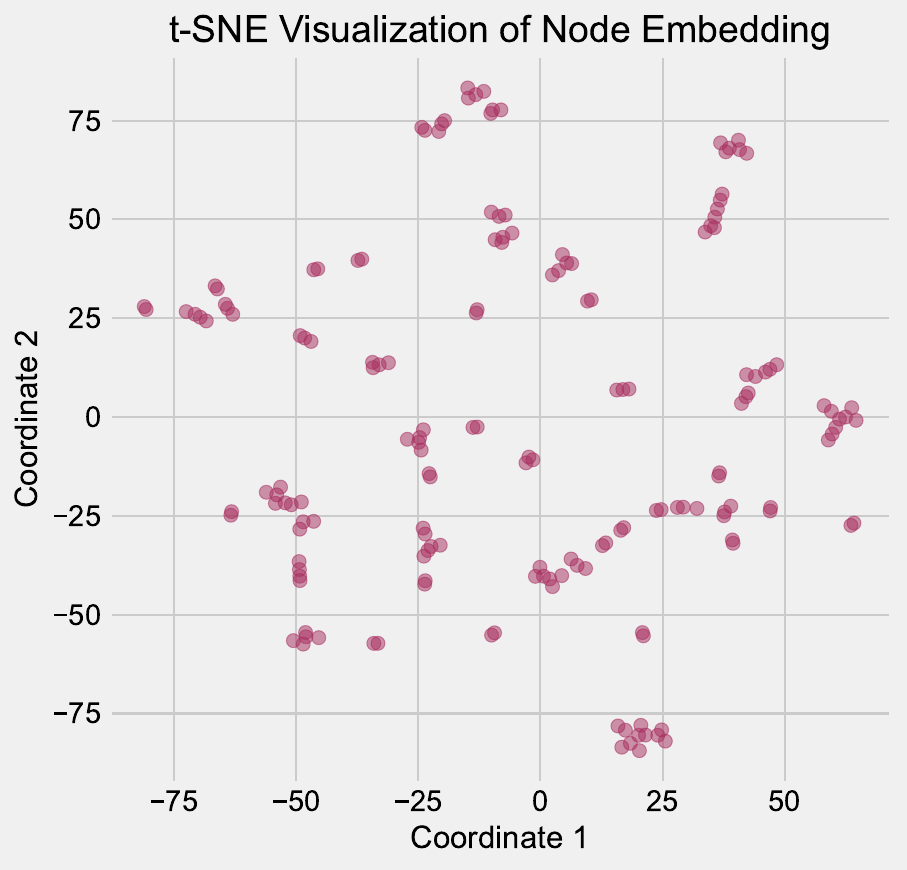}
  \captionsetup{skip=1pt}
  \caption{The t-SNE visualization of node embedding.}
  \label{fig:node_embed_tsne}
\end{minipage}\hfill
\begin{minipage}[c]{0.49\columnwidth}
  \centering
  \includegraphics[scale=0.27]{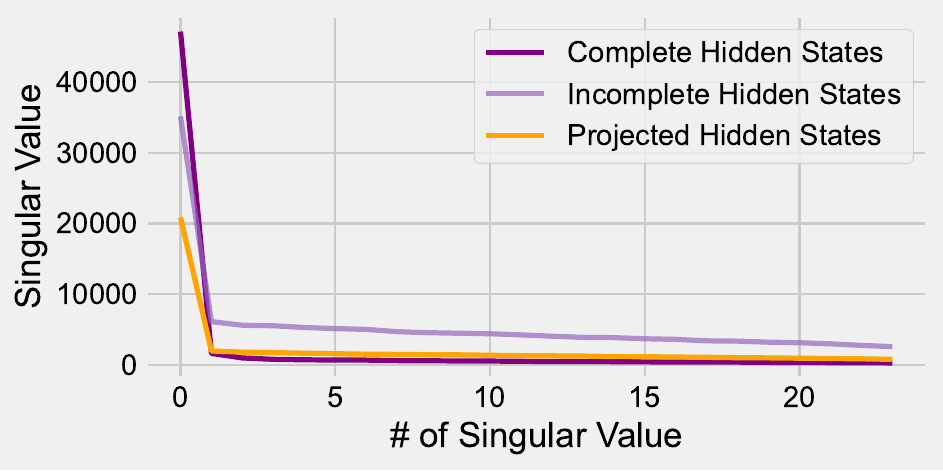}
  \captionsetup{skip=1pt}
  \caption{Singular values of different hidden states in the temporal attention layer.}
  \label{fig:projector_svd}
\end{minipage}
\end{figure}

\noindent\textbf{Interpretations on Spatial Embedding}.
To illustrate the role of node embedding, we analyze the SV spectrum of the \texttt{PEMS08} data in Fig. \ref{fig:node_embed}. The complete data show a prominent low-rank property, but the SVs of incomplete data dramatically expand. In contrast, the node embedding also displays a similar low-rank distribution, which can act as a dense surrogate for each sensor. Furthermore, we analyze the multivariate attention map obtained by correlating the node embedding in Fig. \ref{fig:sptial_att}. 
It is evident that as the embedded attention layers become deeper, the learned attention maps approach the actual ones. However, \textit{incomplete data produce noisy correlations with little informative pattern}. Fig. \ref{fig:node_embed_tsne} displays the t-SNE visualization of each node embedding with two projected coordinates in \texttt{PEMS08}. The embeddings tend to form clusters, and different clusters are apart from others. This phenomenon is in accordance with highway traffic sensor systems that proximal sensors share similar readings.

\begin{figure}[!htbp]
\centering
\includegraphics[scale=0.4]{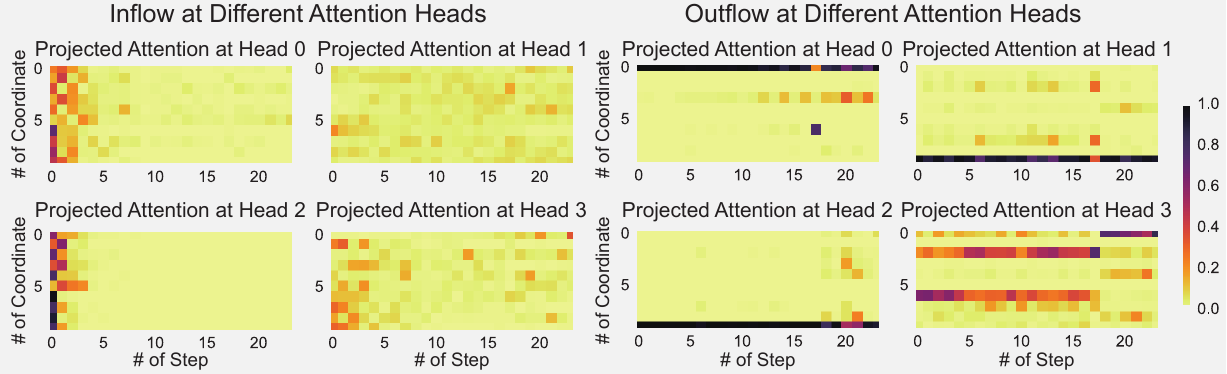}
\captionsetup{skip=1pt}
\caption{Inflow and outflow in the projected attention layer.} 
\label{fig:projector}
\end{figure}

\noindent\textbf{Interpretations on Temporal Projector}. \label{case_study:projector}
To illustrate the mechanism of the temporal projected attention in Eqs. \eqref{proj_att} and \eqref{proj_att_all}, inflow and outflow attention maps are shown in Figs. \ref{fig:projector}. It can be seen that these matrices quantify how the information of incomplete states flows into compact representations and then is recovered to complete states. Inflows show that only a fraction of the message is directed towards the projector, while different attention heads can provide varying levels of information density. Meanwhile, outflows indicate that a small number of temporal modes can reconstruct useful neural representations for imputation. 
\textit{This can be analogous to the low-rank reconstruction process, which serves as an inductive bias for time series with low information density}.
We further examine the SV distribution of different hidden states in the last temporal attention layer. As evidenced by Fig. \ref{fig:projector_svd}, after flow through the projected attention layer, the hidden states have lower SVs than the incomplete inputs and are closer to the complete representations.

\section{Conclusion}\label{conclusion}
This paper demonstrates a low rankness-induced Transformer model termed \texttt{\gls{model}} to address the missing spatiotemporal data imputation problem. Taking advantage of the low-rank factorization, we design projected temporal attention and embedded spatial attention to incorporate structural priors into the Transformer model. Furthermore, a Fourier sparsity loss is developed to regularize the solution's spectrum.
The evaluation results on various benchmarks indicate that \texttt{\gls{model}} not only consistently achieves state-of-the-art imputation accuracy, but also exhibits high computational efficiency, generalizability across various datasets, versatility for different scenarios, and interpretability. Therefore, we believe that it has the potential to advance research on spatiotemporal data for general imputation tasks. Future work can adopt it to achieve time series representation learning task and explore the multipurpose pretraining problem for time series.

\begin{acks}
The work was supported by research grants from the National Natural Science Foundation of China (52125208), National Key R\&D Programs of China (2022YFB2602100), the China National Postdoctoral Program for Innovative Talents (BX20220231), the China Postdoctoral Science Foundation (2022M712409), and the Science and Technology Commission of Shanghai Municipality (22dz1203200). 
\end{acks}

\bibliographystyle{ACM-Reference-Format}
\bibliography{references}

\clearpage
\appendix
\section{Appendix}

\subsection{Additional Discussions}
\subsubsection{Architectural Comparison}
\begin{figure}[!ht]
\centering
\includegraphics[scale=0.7]{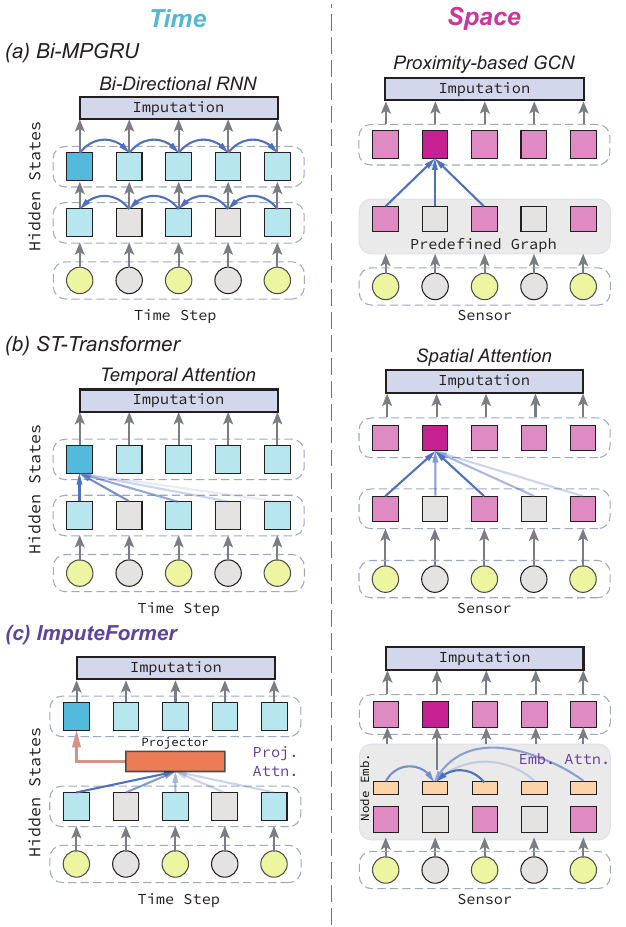}
\captionsetup{skip=1pt}
\caption{(a) MPGRU adopts Bi-RNNs to gather available readings from consecutive time points and GCNs to collect neighborhood data on predefined graphs. (b) Transformers compute all pairwise correlations of the raw data in both spatial and temporal axes. (c) \texttt{\acrshort{model}} utilizes projected attention along the temporal axis and embedded attention based on the node representation in the spatial axis.}
\label{architec_compare}
\end{figure}

We provide an illustration to show the structural difference between different paradigms in Fig. \ref{architec_compare}. 
%

\subsubsection{Effective Input Embedding}\label{appendix: input_embedding}
Unlike images or languages, time series have low semantic densities \cite{nie2022time}. Therefore, many forecasting models flatten and abstract the input series to reduce information redundancy \cite{STID,nie2023contextualizing}. Specifically, given the input $\mathbf{X}_{t:t+T}\in\mathbb{R}^{N\times T}$, each series can be processed by a MLP shared by all series: $\mathbf{z}^{i,(0)}=\texttt{MLP}(\mathbf{x}^{i})$, where $\texttt{MLP}(\cdot):\mathbb{R}^T\rightarrow\mathbb{R}^D$.
However, we claim that this technique is not suitable for imputation. If we express it as follows:
\begin{equation}\label{autoreg}
    x_{t+h}=\sigma\left(\sum_{k=0}^T w_{k,h}x_{t+k}+b_{k,h}\right),~h\in\{0,\dots,T\},
\end{equation}
it is evident that the linear weights only depends on the relative position in the sequence and are agnostic to the data flow.
Since missing data points and intervals can occur at arbitrary locations in the series, fixed weights can learn spurious relationships between each time step, thus overfitting the missing patterns in the training data. 
Therefore, we suggest not to use linear mappings on the time axis to account for the varying missing time points.

\subsubsection{Low-Rankness in Self-Attention}\label{appendix: low-rank}
\citet{wang2020linformer} studied the observation that the self-attention matrix in Transformer is low-rank and proposed a linear attention. Our proposed temporal projected attention model shares a similar idea as this work but has different mechanisms and formulations. We indicate the differences in the following exposition. Given $\mathbf{Q},\mathbf{K},\mathbf{V}\in\mathbb{R}^{T\times D}$, and the projector $\mathbf{P}\in\mathbb{R}^{C\times D}$, temporal projected attention is formulated as:
\begin{equation}\label{eq:appendix_proj_att}
\begin{aligned}
&\texttt{SelfAtten}(\mathbf{Q},\mathbf{P},\texttt{SelfAtten}(\mathbf{P},\mathbf{K},\mathbf{V})), \\
&=\sigma(\mathbf{Q}\mathbf{P}^{\mathsf{T}})\texttt{SelfAtten}(\mathbf{P},\mathbf{K},\mathbf{V})=\underbrace{\sigma(\mathbf{Q}\mathbf{P}^{\mathsf{T}})}_{T\times C}\underbrace{\sigma(\mathbf{P}\mathbf{K}^{\mathsf{T}})}_{C\times T}\mathbf{V},
\end{aligned}
\end{equation}
while the linear attention \cite{wang2020linformer} assigns two learnable matrices $\mathbf{E},\mathbf{F}\in\mathbb{R}^{C\times T}$ and has the following form:
\begin{equation}\label{eq:appendix_linear_att}
\texttt{SelfAtten}(\mathbf{Q},\mathbf{E}\mathbf{K},\mathbf{F}\mathbf{V})), \\
=\sigma(\underbrace{\mathbf{Q}\mathbf{K}^{\mathsf{T}}}_{T\times T}\underbrace{\mathbf{E}^{\mathsf{T}}}_{T\times C})\mathbf{F}\mathbf{V}.
\end{equation}

As for complexity, we can compute Eq. \eqref{eq:appendix_proj_att} in the order: $\sigma(\mathbf{QP}^{\mathsf{T}})>\sigma(\mathbf{PK}^{\mathsf{T}})>\sigma(\mathbf{PK}^{\mathsf{T}})\mathbf{V}>\sigma(\mathbf{QP}^{\mathsf{T}})\sigma(\mathbf{PK}^{\mathsf{T}})\mathbf{V}$, which admits $O(4TDC)$ complexity. Similarly, Eq. \eqref{eq:appendix_linear_att} has the same complexity.
Although of the same complexity, our model has an explicit and symmetric formulation that brings improved model expressivity. The advantages are threefold: 
(1) \textit{explicit low-rank factorization}: Linformer \cite{wang2020linformer} does not directly achieve a low-rank factorization of the attention matrix. It first computes the full attention matrix $\mathbf{QK}^{\mathsf{T}}$ with size $T\times T$ and then compresses it to $T\times C$. Instead, \texttt{ImputeFormer} directly factorizes the full attention matrix from $T\times T$ to $T\times C$, $C\times T$, which is beneficial for dealing with redundancy and missingness in the attention matrix; 
(2) \textit{pattern adaptation}: Linformer sets the compression matrix $\mathbf{E},\mathbf{F}\in \mathbb{R}^{C\times T}$ completely learnable, which is agnostic to the missing patterns of $\mathbf{Q}$, $\mathbf{K}$, and $\mathbf{V}$. These static parameters cannot account for the varying missing patterns. Instead, we obtain the $T\times C$ factor matrix through the query $\mathbf{QP}^{\mathsf{T}}$, which is pattern-adaptive; 
(3) \textit{increased capacity}: Linformer has $T\times C$ learnable parameters, while ours has $C\times D$ parameters, which has a larger model capacity while having the same time complexity.

\subsection{Reproducibility}\label{Appendix:settings}
\subsubsection{Implementations}

\begin{figure*}[!htbp]
\centering
\begin{subfigure}[b]{0.33\textwidth}
\centering
\includegraphics[scale=0.36]{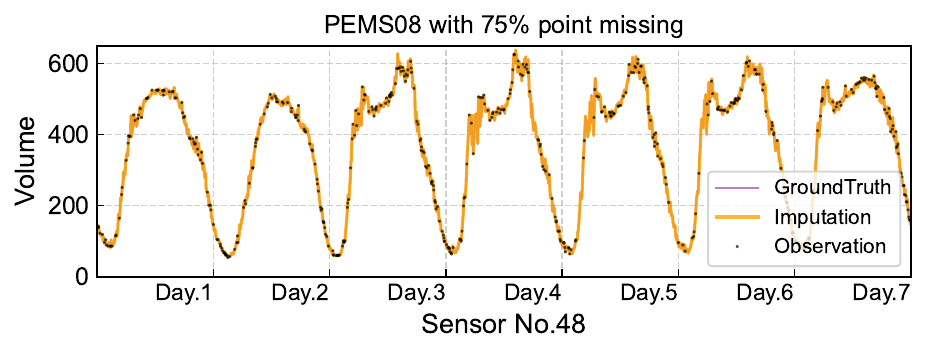}
\captionsetup{skip=1pt}
\caption{PEMS08, $75\%$ point missing}
\end{subfigure}
\begin{subfigure}[b]{0.33\textwidth}
\centering
\includegraphics[scale=0.36]{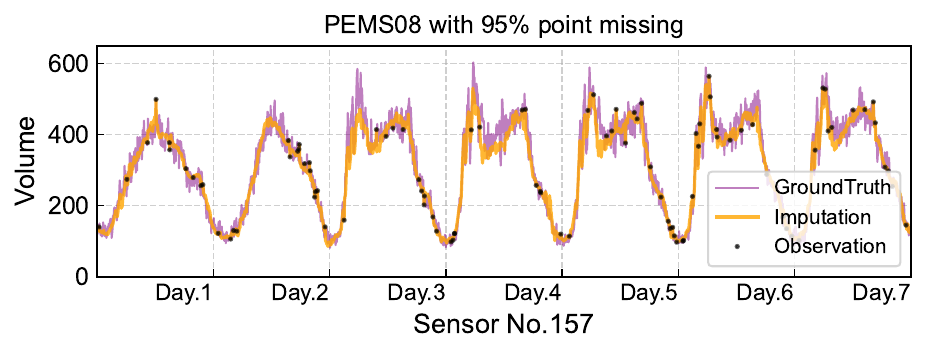}
\captionsetup{skip=1pt}
\caption{PEMS08, $95\%$ point missing}
\end{subfigure}
\begin{subfigure}[b]{0.33\textwidth}
\centering
\includegraphics[scale=0.36]{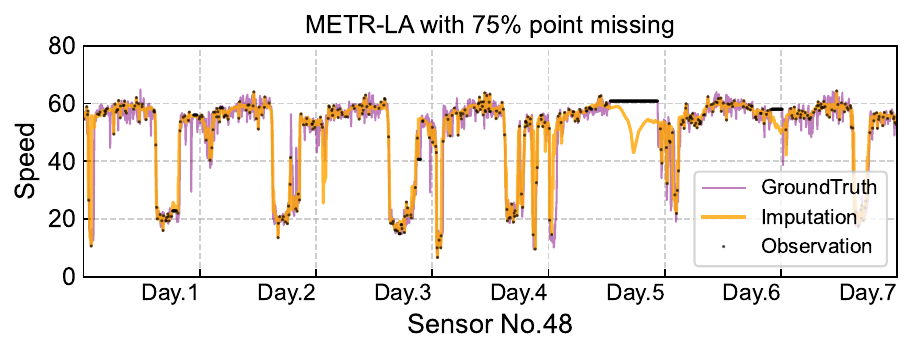}
\captionsetup{skip=1pt}
\caption{METR-LA, $75\%$ point missing}
\end{subfigure}
\begin{subfigure}[b]{0.33\textwidth}
\centering
\includegraphics[scale=0.36]{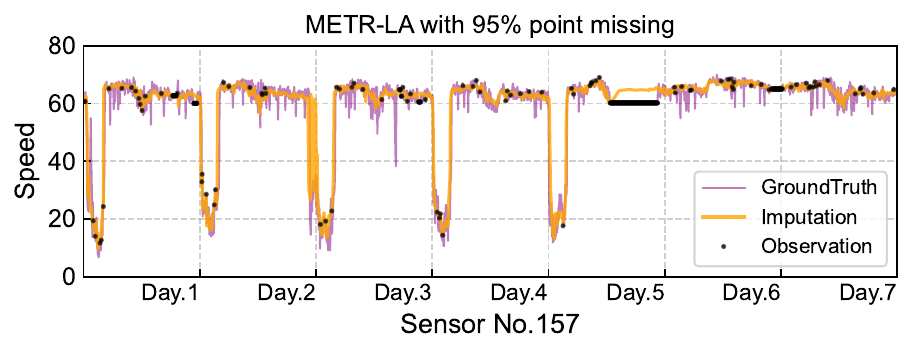}
\captionsetup{skip=1pt}
\caption{METR-LA, $95\%$ point missing}
\end{subfigure}
\begin{subfigure}[b]{0.33\textwidth}
\centering
\includegraphics[scale=0.36]{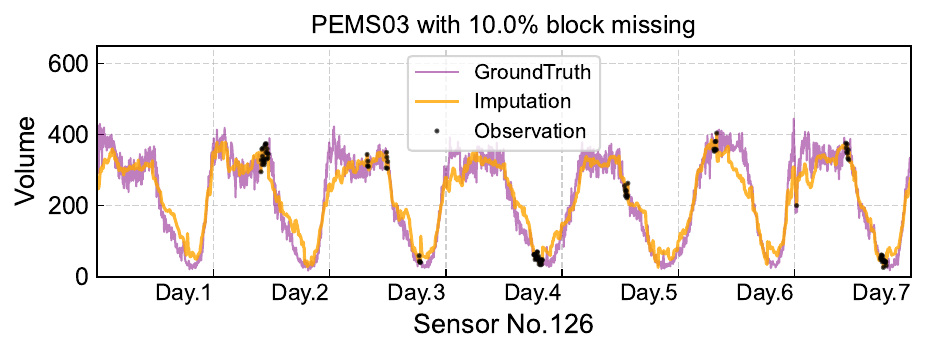}
\captionsetup{skip=1pt}
\caption{PEMS03, $10\%$ block missing}
\end{subfigure}
\begin{subfigure}[b]{0.33\textwidth}
\centering
\includegraphics[scale=0.36]{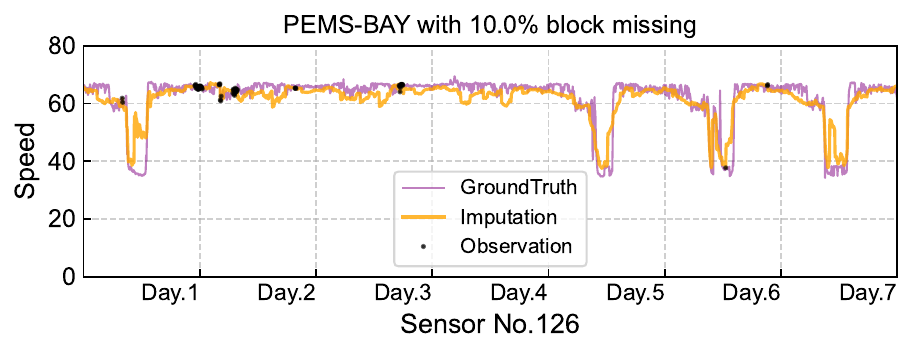}
\captionsetup{skip=1pt}
\caption{PEMS-BAY, $10\%$ block missing}
\end{subfigure}
\captionsetup{skip=1pt}
\caption{Visualization examples of imputation for traffic speed and volume data under different missing patterns.}
\label{fig:example}
\end{figure*}

We build our model and baselines based on the \texttt{SPIN} repository (\url{https://github.com/Graph-Machine-Learning-Group/spin}). All experiments were performed on a single NVIDIA RTX A6000 GPU (48 GB). For the hyperparameters of \texttt{\acrshort{model}}, we set the hidden size to 256, the input projection size to 32, the node embedding size to 64, the projected size to 6, the number of attention layers to 3, and the sequence length to 24 for all data. We also keep the same training, validation and evaluation split as \cite{GRIN,SPIN} and report the metrics on the masked evaluation points.

\subsubsection{Dataset Descriptions}
We adopt heterogeneous spatiotemporal benchmark datasets to evaluate the imputation performance. 

\noindent\textbf{{Traffic Speed Data}}. Our experiments include two commonly used traffic speed datasets, named \texttt{METR-LA} and \texttt{PEMS-BAY}. \texttt{METR-LA} contains spot speed data from 207 loop sensors over a period of 4 months from Mar 2012 to Jun 2012, located at the Los Angeles County highway network. \texttt{PEMS-BAY} records 6 months of speed data from $325$ static sensors in the San Francisco South Bay Area.

\noindent\textbf{{Traffic Volume Data}}. We adopt four traffic volume data, including \texttt{PEMS03}, \texttt{PEMS04}, \texttt{PEMS07}, and \texttt{PEMS08}. They contain the highway traffic volume record collected by the Caltrans Performance Measurement System (PeMS) and aggregated into 5-minute intervals. 

\noindent\textbf{{Energy and Environmental Data}}. Four energy and environmental data are selected to evaluate the generality of models, including: (1) \texttt{Solar}: solar power production records from 137 synthetic PV farms in Alabama state in 2006, which are sampled every 10 minutes; (2) \texttt{CER-EN}: smart meters measuring energy consumption from the Irish Commission for Energy Regulation Smart Metering Project \cite{cer2012}. Following the setting in \cite{GRIN}, we select 435 time series aggregated at 30 minutes for evaluation.
(3) \texttt{AQI}: PM2.5 pollutant records collected by 437 air quality monitoring stations in 43 Chinese cities from May 2014 to April 2015 with the aggregation interval of 1 hour. Note that \texttt{AQI} data contains nearly $26\%$ missing data. (4) \texttt{AQI36}: a subset of \texttt{AQI} data which contains 36 sensors in Beijing distinct. 

\subsubsection{Experimental Settings and Baseline Methods}
This section describes the detailed information on experimental setups.

\noindent\textbf{{Missing patterns}}. For traffic, {Solar} and {CER-EN}, we consider two scenarios discussed in \cite{GRIN,SPIN}: (1) Point missing: randomly remove observed points with $25\%$ probability; (2) Block missing: randomly drop $5\%$ of the available data and at the same time simulate a sensor failure lasting for $\mathcal{L}\sim \mathcal{U}(12,48)$ steps with $0.15\%$ probability. We keep the above missing rates the same as in the previous work \cite{GRIN,SPIN}. In addition, we also evaluated the performance under sparser conditions. For example, the block missing with $10\%$ probability corresponds to a total missing rate of $\approx 90\sim 95\%$. Note that matrix or tensor models can only handle in-sample imputation, where the observed training data and the test data are in the same time period. However, deep models can work in out-of-sample scenarios \cite{GRIN} where the training and test sequences are disjoint. We adopt the out-of-sample tests for deep models and in-sample tests for others.

\noindent\textbf{{Baseline Methods}}. We compare our model with SOTA deep-learning and low-rank imputation methods. For statistical and optimization models, we consider: (1) Observation average (\texttt{Average}); (2) Temporal regularized matrix factorization (\texttt{TRMF}) \cite{yu2016temporal}; (3) Low-rank autoregressive tensor completion (\texttt{LRTC-AR}) \cite{chen2021low}; (4) \texttt{MICE} \cite{van2011mice}. For deep imputation models, we select several competitive baselines: (1) \texttt{SPIN} \cite{SPIN}: sparse spatiotemporal attention model with state-of-the-art imputation performance; (2) \texttt{GRIN} \cite{GRIN}: message-passing-based bidirectional RNN model with competitive performance; (3) \texttt{SAITS} \cite{du2023saits}: Temporal Transformer model with diagonally masked attention; (4) \texttt{BRITS} \cite{cao2018brits}: bidirectional RNN model for imputation; (5) \texttt{rGAIN} \cite{yoon2018gain}: GAIN model with bidirectional recurrent encoder and decoder; (6) \texttt{Transformer}/\texttt{ST-Transformer} \cite{transformer}: canonical Transformer with self-attention in temporal or spatial-temporal dimensions; (7) \texttt{TiDER} \cite{liu2022multivariate}: matrix factorization with disentangled neural representations; (8) \texttt{TimesNet} \cite{wu2022timesnet}: 2D convolution-based general time series analysis model; (9) \texttt{BiMPGRU}: a bidirectional RNN-based GCN model, which is similar to DCRNN \cite{li2017diffusion}.

\noindent\textbf{Ablation Studies}.
Particularly, we examine the following variations: (a) Temporal blocks: we replace the temporal interaction module with $\texttt{MLP}$ or directly remove it; (b) Spatial blocks: we replace the spatial interaction module with $\texttt{MLP}$ or directly remove it; (c) Loss function: We remove the FIL or replace it with a hierarchical loss used in \cite{SPIN,du2023saits}; (d) Architecture: we evaluate the impacts of the order of spatial-temporal blocks and the joint attention strategy. 
We adopt two data to evaluate and other settings remain the same. 

\noindent\textbf{Evaluation Metrics}.
To evaluate the model, we simulate different observation conditions by removing parts of the raw data to construct incomplete samples based on different missing rates ($p_{\text{missing}}$). Evaluation metrics are then calculated for these simulated missing points. We use a masking indicator $\mathbf{M}_{\text{missing}}$ to denote these locations in which the unobserved (missing) values are marked as ones, observed as zeros. Note that the masked points for evaluation are not available for the models during all stages. The mean absolute error (MAE) is adopted to report the results.

\subsection{Imputation Visualization}
We provide several visualization examples in Fig. \ref{fig:example}. As evidenced, \texttt{\acrshort{model}} can generate reasonable imputations by learning the inherent structures of spatiotemporal data. Previous studies have discovered that low-rank models can cause oversmoothing estimation \cite{chen2021low,nie2023correlating}. Due to the representation power of deep architectures, our model can provide a detailed reconstruction. In particular, although only limited temporal information is available in the block missing case, it can resort to the node embedding as the query to spatial relations, thereby generating an effective imputation.

\end{document}